\documentclass[BCOR=1mm, DIV=calc,10pt,
twoside=true,
twocolumn,
headings=normal]{scrartcl}
\usepackage{microtype}
\usepackage{graphicx}
\usepackage{booktabs} % for professional tables

\usepackage{amsmath}
\usepackage{bm}
\usepackage{url}

\usepackage{cite}
\usepackage{amsmath,amssymb,amsfonts}
\usepackage{graphicx}
\usepackage{textcomp}
\usepackage{xcolor}

\usepackage{authblk}

\newcommand{\fig}{Fig.~}
\newcommand{\eqn}{Eqn.~}
\newcommand{\tab}{Tab.~}

\newcommand{\etal}{ {\em et al.}}

\begin{document}

\title{Demand Forecasting of Individual Probability Density Functions with Machine Learning}

% Michael am Ende als "PI"
\author[1]{Felix Wick}
\author[3]{Ulrich Kerzel}
\author[1]{Martin Hahn}
\author[1]{Moritz Wolf}
\author[2]{Trapti Singhal}
\author[1]{Daniel Stemmer}
\author[1]{Jakob Ernst}
\author[1]{Michael Feindt}
\affil[1]{\small Blue Yonder GmbH (Ohiostra{\ss}e 8, 76149 Karlsruhe, Germany)}
\affil[2]{\small Blue Yonder India Private Limited (Bengaluru, India)}
\affil[3]{\small IU Internationale Hochschule (Erfurt, Germany)}

\date{}

\maketitle

\begin{abstract}

Demand forecasting is a central component of the replenishment process for retailers, as it provides crucial input for subsequent decision making like ordering processes. In contrast to point estimates, such as the conditional mean of the underlying probability distribution, or confidence intervals, forecasting complete probability density functions allows to investigate the impact on operational metrics, which are important to define the business strategy, over the full range of the expected demand. Whereas metrics evaluating point estimates are widely used, methods for assessing the accuracy of predicted distributions are rare, and this work proposes new techniques for both qualitative and quantitative evaluation methods. Using the supervised machine learning method ``Cyclic Boosting'', complete individual probability density functions can be predicted such that each prediction is fully explainable. This is of particular importance for practitioners, as it allows to avoid ``black-box'' models and understand the contributing factors for each individual prediction. Another crucial aspect in terms of both explainability and generalizability of demand forecasting methods is the limitation of the influence of temporal confounding, which is prevalent in most state of the art approaches.

\end{abstract}


{DOI: 10.1007/s43069-021-00079-8}

{Published in Springer Nature Operations Research Forum.}

\section{Introduction}
\label{sec:Intro}

% specific quantile
One of the main business operations of retailers is to ensure that sufficient goods are available in the store for customers to buy. This entails two parts: In a first step, we need to estimate the future customer demand to be able to judge how many goods we will need to order from the supplier or wholesaler, given the current inventory level. In the second step, aimed at inventory control, we need to place the actual orders, which is further complicated by the fact that multiple operational constraints, such as lot-sizes or delivery schedules, need to be included in the final ordering decision. Unfortunately, from the perspective of the retailer, the two parts, estimating the future demand and optimization of the order quantity, do not factorize into two separate issues, but are inextricably linked. This is because the business strategy depends on the complete chain of demand estimation and subsequent replenishment process to control the inventory. We can understand this in the following way: In the simplest case, a retailer might want to optimize their profit, or equivalently the cost, by reducing the impact of perishable items that have not been sold at the end of the selling period, as well as the lost revenue by not being able to meet the customer demand within this period. This is known as the ``newsvendor problem'' \cite{Edgeworth}, see, e.g., \cite{Khouja1999537} for a detailed review. Assuming linear overage ($h$) and underage ($b$) cost, the optimal quantile $q_{\mathrm{opt}} = {b}/{(b+h})$ of a known probability density function (PDF) describing demand as random variable can be calculated exactly. This quantile then corresponds to the numerical value of the demand PDF prediction that can be used to determine the optimal order quantity.

% simulations
In reality, the situation is often more complex, as many factors may contribute to the cost function. Furthermore, the cost function may vary between individual stock-keeping units (SKU) and may also change with time. However, the general principle remains: If we have access to the relevant cost function, we can determine the optimal point estimator of the future demand from the associated PDF, see Appendix \ref{sec:CostQuantile}. For the case of a relatively simple cost function, such as the mean squared error (MSE) or the mean absolute deviation (MAD), the calculations are straight-forward. If the cost function becomes more complex, we may no longer be able to perform the calculations analytically, but will have to resort to numerical approaches. Furthermore, if we can predict more than one quantile, we can investigate the effect of choosing different quantiles, and determine, for example, how sensitive operational metrics, such as stockout or waste rate of perishable goods, change. This can be done using, for example, a business-oriented simulation that starts from the quantiles of the predicted demand for each SKU to simulate the value of the business-relevant KPIs for each setting, taking operational constraints into account. This can then be used, for example, to align the business strategy. Predicting the full demand PDF therefore allows the maximum flexibility for further optimizations, because we have direct access to all quantiles.

% PDF predictions
Due to its stochastic nature, demand is difficult to forecast: It depends on many influencing factors and, as already stated above, we can interpret it as a random variable with associated probability distribution. And as such, demand is neither identically and independently distributed (i.i.d.) between different products and sales locations, nor between different times. This means that, while we can generally assume that demand follows a specific type of PDF, its parameters are unique to the instance for which an estimate is required. For example, the PDF describing the demand of a particular product-location-date combination is specific to the product, date, and retail location for which the forecast is made. Furthermore, this distribution depends on a wide range of influencing factors, such as seasonal effects, sales price, local weather, etc. Therefore, in order to predict the expected demand for each SKU at each sales location and sales day, we need to forecast individual probability distributions rather than globally adjusted parameters for groups of SKUs, locations, or sales days.

% temporal confounding
If we account for the cases where demand exceeds the current stock level, resulting in out-of-stock situations, the actual demand in the past can be approximated by realized sales. We can then understand the time-ordered observations of sales events as a time series of a random variable. It is therefore tempting to use established time series regression models to predict the future values of the expected demand. However, if we wish to use the predicted values for anything but the prediction itself, we have to be careful to avoid, or at least limit as much as possible, temporal confounding, which emerges if some external variable $X$ causally influences several time steps $t$ of the observed time series variable $Y$, say $Y_{t}$ and $Y_{t-\mathrm{lag}}$, where ``lag'' denotes a specific time difference between the two steps. In the case of a retailer, $Y_{t}$ is the demand for a specific product at a given sales location for a specific day, and $X$ can be, for example, the price of the product, which we can set as part of our business strategy or change according to some promotion. This dependency creates a backdoor path between $Y_{t-\mathrm{lag}}$ and $Y_{t}$, that in turn leads to a spurious autocorrelation between the values of $Y$ at different times $t$, if we cannot control for the confounder $X$. This is particularly important for the case of demand forecasting, since customer demand is not a natural phenomenon we observe as an outsider, but retailers actively influence the customer demand  through a wide range of interventions, such as a given assortment choice, promotions, advertisements, coupons, etc. If we cannot disentangle the fundamental causal dependencies from the spurious autocorrelation, understanding the influences that lead to a specific prediction becomes very challenging, even if the underlying prediction method itself is fully explainable. Moreover, due to this disguising of actual causal dependencies, reliance on temporal confounding is also detrimental in terms of generalizability of the forecasting model. The issue of temporal confounding, together with a proposed solution, is discussed in more detail in sec. \ref{sec:ts_confounding}.

% ML
Demand forecasting for retailers often requires to predict millions of product-location combinations each day. Univariate time series forecasting methods operate on each of these many individual time series separately. On the other hand, supervised machine learning (ML) algorithms optimize all individual time series simultaneously, which can enhance the generalizability of the method by exploiting common aspects between the various time series. This can lead to a drastic reduction in variance of the model, and in turn to an improvement of the forecast quality of each of the individual time series. Furthermore, machine learning provides a straight-forward way to include the many exogenous variables that influence demand as features (or covariates) of the model. This can also help to improve the forecast quality compared to methods mainly relying on autocorrelation of the historic sales.

Hence, in order for a retailer to be able to use demand forecasts most effectively, we need to be able to:
\begin{itemize}
	\item {Predict the future demands on SKU level at required temporal granularity as complete individual PDFs.}
	\item {Avoid temporal confounding in the model used for the prediction of the future demand.}
	\item {Evaluate that each part of the PDF models the observed data appropriately.}
	\item {Ideally, each prediction is fully explainable, enabling operational experts to exactly understand what contributed to it.}
\end{itemize}

The remainder of the paper is organized as follows: We first review the relevant literature and existing work in sec. \ref{sec:LitRev}. We then describe our method to predict individual negative binomial PDFs using  a parametric approach including two distinct machine learning models for mean and variance in sec. \ref{sec:pdfEstimation}. After that, we describe novel techniques for the qualitative and quantitative evaluation of PDF predictions in sec. \ref{sec:pdfEvaluation}. This is followed by a detailed discussion on temporal confounding in sec. \ref{sec:ts_confounding}. And finally, we present a demand forecasting example to show an application of our methods in sec. \ref{sec:example}.

\subsection*{Summary of Contributions}

This work brings together elements from three different fields: operations research, machine learning, and causal inference. The combination of these allows to build state-of-the-art demand forecasting models that are fully explainable and incorporate the causal structure of demand. The contribution of this paper is twofold:

Using the explainable machine learning algorithm Cyclic Boosting \cite{Wick2019}, we show how complete probability distributions of the future demand of individual products in retail stores can be predicted at the operationally required granularity. In particular, we show how using machine learning, the detrimental effects of temporal confounding, that are intrinsic to many methods used for time series forecasting, can be limited. The impact of temporal confounding in the context of time series forecasting has not been widely investigated so far in the scientific literature, and to the best of our knowledge this is the first discussion within the context of demand forecasting.

Furthermore, we show how predictions of full individual probability distributions, no matter if created by an approach with or without assumption of a specific distribution class, can be evaluated qualitatively and quantitatively, such that we can assess whether both overall functional form (in the case of a parametric approach) as well as specific quantiles are predicted correctly. This allows to verify that the predicted demand distribution appropriately reflects the observed data in terms of model choice and forecast accuracy.

The general methods developed in this paper can be applied to any setting that requires the prediction and evaluation of probability distributions from observational data, not just demand forecasting. However, to make the impact and benefits more tangible in a practical setting, as well as describe a general methodology how to build such systems from a practitioner's perspective, we limit the discussion to demand forecasting in retail.

\section{Literature Review}
\label{sec:LitRev}

Demand forecasting is one of the core operational activities of any customer-facing business and accurate demand predictions are vital to the optimization of the business operations. This is of particular concern, for example, for supermarkets, as their profit margin tend to be very low, e.g., 1\% and lower for European discounters \cite{MarginLidlSweden} to about 4\% for regular supermarkets \cite{MarginWillysSweden,MarginHemkopSweden} to about 8\% in the USA \cite{MarginPublixUSA}. It can be approached in a variety of ways, see, e.g., \cite{beheshti2015survey} for a survey of methods used in retail sales forecasting with a particular focus on fashion retail.

% univariate time series methods
A popular approach is to treat the demand as a time series, i.e., as a time-ordered sequence of observed sales events. In this setting, the future demand can be predicted as a point estimate using auto-regressive integrated moving average (ARIMA) \cite{BoxJenkins} or exponential smoothing \cite{croston1972forecasting, holt1957forecasting, Brown1963, Gardner1985} approaches. See also \cite{deGooijerHyndman2006} for a comprehensive overview over traditional time series forecasting methods. Essentially, these methods extrapolate the future values by building some form of regression model based on past observations, i.e., $y_{t+1} = f(y_t, y_{t-1}, \ldots)$, introducing dedicated functional components for trends, seasonality, or external factors in various ways, depending on the exact modeling approach taken. The main underlying concept behind these approaches is to exploit the autocorrelation between values of the variable $Y$ at different times $t$. Intuitively, we assume that  the sequence of events observed so far is a good predictor of future events. Examples for these methods in the context of demand forecasting can be found in, e.g., \cite{huber2017cluster, kalchschmidt2006forecasting, fattah2018forecasting, permatasari2018sales}.  A different approach is to use a concept from engineering, the Kalman filter \cite{kalman1960new}, and apply it to statistical forecasting \cite{morrison1977kalman}. Examples for this method used to estimate demand are, e.g., \cite{mitropoulos1980using, tegene1991kalman, kandananond2014applying,jacobi2007water}. Starting from the concept of Kalman filters, state space models \cite{hyndman2008forecasting} relate the measured values over time to the evolution of (unknown) states of the considered system. A comparison of the performance of ARIMA and state space models in a retail setting can be found, e.g., in \cite{ramos2015performance}. Structural time series models \cite{StructuralTS} take a similar approach and decompose the observations into components for trend, seasonality, and others. A modern implementation is Facebook's ``Prophet''  \cite{taylor2018forecasting}. Several other approaches are known in the literature as well: K\"ok and Fisher \cite{kok2007demand} model the demand based on the average demand for a product, modified by a logistic regression for the number of customers in a store, purchase incidence, and product choice. Wang \etal \cite{wang2005demand} model the demand under the assumption of a Poisson process.

% ML
Alternatively to the methods discussed above, time series forecasts in general and demand predictions in particular can be approached by means of supervised machine learning methods, using the values of the time series to be predicted as target, i.e., the numbers the machine learning algorithm is trained to predict.  For example, see \cite{Zhang2012,remus2001neural} for applications of neural networks for time series forecasting. Machine learning approaches can include multiple covariates. This allows to include not only past sales data, but also information about prices, promotions, or details about the article or store in the forecast. There are two main approaches for time series forecasting using  machine learning: We either assume that the variables are independent random variables (for both prediction target and potentially time-dependent features) for each time step of the different time series, or we consider the full sequences of time steps as a whole, resulting in a single sample (including prediction target and potentially time-dependent features) for each individual time series. For the case of individual samples for different time steps, in order to combine the endogenous information from the target autocorrelation with all other exogenous features, the most prominent approach is to include lagged target information explicitly via stacking of univariate time series predictions for the target as features in the machine learning model. For example, we could add exponentially smoothed moving averages with a range of damping factors as features into the model. An example for the estimation of future demand in the context of price optimization, using regression trees \cite{breiman1984}, can be found in \cite{ferreira2016analytics}. For the case of sequence samples, deep learning methods such as recurrent neural networks (RNN) \cite{rnn}, especially in form of ``long short-term memory'' (LSTM) networks \cite{hochreiter1997long}, or transformers \cite{transformer}, based on the self-attention mechanism, are used. These methods employ information about the prediction target autocorrelation implicitly via the input sequences of the time-ordered training events. Both concepts, LSTM cells and self-attention, aim at learning  the elements of the individual time series from the data that are deemed important for the prediction of the future values by the algorithm. The interest in deep learning approaches for demand forecasting has increased significantly recently, see, e.g., \cite{bandara2019sales,yu2017application, goyal2018solution,helmini2019sales, golkabek2020demand}. A review of deep learning methods for time series forecasting in a wide range of applications can also be found in \cite{langkvist2014review}, and the use of RNNs with focus on industrial applications is discussed in \cite{dixon2020industrial}. A further approach is to explore the use of generative adversarial networks \cite{goodfellow2014generative} in the context of time series forecasting \cite{haas2020statistical, ramponi2018t, smith2020conditional}. 

% temporal confounding
In many situations, relying on the exploitation of autocorrelation, either via traditional time series forecasting or machine learning approaches, is sufficient. However, this can lead to shortcomings in terms of explainability and generalizability due to temporal confounding. There are two main paths to avoid this: One way is to explicitly account for the confounders and extend the model to include them. The main challenge is then the discovery of temporal confounders and, for example, try to learn the causal structure from the observational data, see, e.g., \cite{malinsky2018causal,runge2018causal,Runge2019}. This topic is of particular importance in fields like medicine \cite{bica2020time} or climate research \cite{perrakis2014controlling}. Brodersen \etal \cite{Brodersen2015,GoogleCausalImpact} propose to use Bayesian structural time series models to estimate the causal effect of interventions on a time series by creating a counterfactual prediction that describes the behavior of the time series had the intervention not taken place. While structural time series equations allow to model components such as trend, seasonality, and others explicitly, the fundamental form of the state equation is $Y_t = G_t Y_{t-1} + \epsilon_t$, where $\epsilon_t \sim \mathcal{N}(0,\sigma^2)$ is a white noise term and $G_t$ some matrix that describes the evolution from the state $Y_{t-1}$ to $Y_t$. Therefore, great care must be taken to avoid temporal confounding when setting up the states for structural time series models. Compared to fields like medicine, marketing, and many others that focus on the causal impact of few or just one intervention(s), retailers have to include many external effects that affect the predicted demand and are repeated frequently: Prices are changed dynamically, advertisements placed, new products are included in the assortment and others removed, etc.

\noindent
The alternative to the above approaches is to separate the autocorrelation from exogenous information via a chain of independent models (see sec. \ref{sec:ts_confounding}). 

% PDF predictions
Common to all approaches discussed so far is that they predict a point estimate, essentially a single number that indicates the next value in the sequence, for example, the number of a specific product likely to be sold in a given store on a particular opening day. While operational decisions can be based on such a single number, this does not allow us to evaluate the uncertainty on such a prediction. Since the model parameters are not {\em a priori} known but estimated from data, the predicted values are generally associated with an uncertainty that needs to be quantified. Several methods have been developed, and Chatfield \cite{chatfield1993calculating} reviews the methods available at the time. In particular, bootstrapping methods for auto-regressive time series forecasting have been proposed, see, e.g.,\cite{masarotto1990bootstrap,mccullough1994bootstrapping,mccullough1996consistent,grigoletto1998bootstrap,thombs1990bootstrap,clements2001bootstrapping,Angus1994,pascual2004bootstrap,pascual2001effects,pascual2005bootstrap}. However, as already pointed out in \cite{chatfield1993calculating}, estimating the uncertainty of the point forecast is only one reason to go beyond a point estimate. For example, if we want to be able to evaluate strategies for a range of different outcomes, we need access to more information. In particular, if we do not wish to make the assumption that the point forecast is the mean of an underlying Gaussian distribution, that can be fully specified by the point forecast and an interval, we need to be able to compute any quantile of the distribution or estimate the full PDF directly.

% quantile regression
Generally, quantile regression \cite{koenker2001} can be implemented in various frameworks and used to estimate a range of quantiles for each predicted distribution, from which an empirical probability distribution can be interpolated. Using a dedicated neural network \cite{Feindt2006190}, either the full PDF or a defined range of quantiles can be calculated directly from the data for each individual prediction without assuming an underlying model. Focusing on time series, Hyndman \cite{hyndman1995highest} proposes to use either a simulation or bootstrapping framework to estimate the highest density forecast regions of non-normal time series. Tay \cite{tay2000density} summarizes the use of density forecasts in applications in finance and economics. Using RNNs, quantile regression for multiple forecasting horizons \cite{wen2017multi} has been used to predict the demand at the internet retailer Amazon. Using an attention-based architecture, the Temporal Fusion Transformer (TFT) \cite{lim2019temporal} can also predict a set of specified quantiles at each predicted time step. Another possibility to model the underlying data distribution without assumption of a fixed distribution class is the usage of conditional normalizing flows \cite{rezende2016variational,timeseriesflow}, which aim to transform a simple initial density into a more complex one by applying a sequence of invertible transformations.

 % PDF assumption
Alternatively, we can start from a model assumption for the demand distribution and fit the model parameters instead of reconstructing the complete distribution \cite{astonpr373, SALINAS20201181}. This approach is computationally favorable and usually more robust, as fewer parameters need to be estimated. Empirically, we can determine the best fitting distribution from data \cite{adan1995}. However, given the stochastic nature of demand, such an empirically determined distribution is not expected to be stable and prone to sudden changes. Instead, the choice of the demand distribution should be motivated by theoretic considerations. Discrete demand is typically modeled as a negative binomial distribution (NBD), also known as Gamma-Poisson distribution \cite{Ehrenberg1959, Ehrenberg1967, Ehrenberg1972, Chatfield1973, Schmittlein_1985}. This distribution arises if the Poisson parameter $\mu$ is a random variable itself, which follows a Gamma distribution. The NBD has two parameters, the mean $\mu$ and the variance $ \sigma^2 > \mu$, and is over-dispersed compared to the Poisson distribution, for which $\sigma^2 = \mu$. Hence, for each ordering decision, the model parameters $\mu$ and $\sigma^2$ need to be determined at the required granularity, typically for each item, sales location, and ordering time, depending on all auxiliary data describing article details, retail location, and influencing factors such as pricing and advertisement information.

% order optimization
Finally, since from an operational perspective the retailer's focus is less on the demand prediction {\em per se} but on the decisions, in particular, the ordering decisions, that are derived from these, we need to consider the question if we should skip the separate step of estimating the demand and directly predict the optimal order quantity. This direct approach is often referred to as ``data-driven newsvendor'' in the recent literature, see, e.g., \cite{beutel2012safety, ban2019big, bertsimas2020predictive, oroojlooyjadid2020applying,huber2019data}. It aims to avoid estimating the underlying PDF for demand and use the available data (historic sales records and further variables)  to derive the operational decisions (i.e., the order quantity) directly. Although the integrated approach seems preferable at first glance, since it avoids determining the full demand distribution and results directly in the desired operational decision, the indirect approach via demand forecasts offers some substantial advantages. First, demand forecasts in form of full PDFs can be used to simulate the performance of the relevant metrics on the level of individual items, and, for example, optimize the impact on business strategy decisions on conflicting metrics such as out-of-stock (i.e., lost sales) and waste rate. From a practitioners perspective, separating the demand forecast from the operational decisions (i.e., calculating the order quantities for the next delivery cycle) enables longer-term planning and reduces the complexity, as it avoids coupling delivery schedules of multiple wholesalers and manufacturers with the forecast of customer demand. It also allows to share long-term demand predictions with other business units or external vendors and wholesalers to ease their planning for the production and supply chain processes upstream of the retailer. From the perspective of the industrial practice of a vendor of supply chain methods and tools, modeling the demand separately from deriving the subsequent orders has the additional benefit that multiple retail chains can benefit from any improvement in the model description, even if the specific retailers are unrelated to each other. Additionally, a purely data-driven approach going from the observed data directly to the operational decision (such as the order quantity) does not allow to analyze the data-generating process, i.e., the mechanism behind the stochastic behavior of the customer demand. This is crucial, for example, if a causal analysis is planned, such as a study of the effect of promotions, advertisements, price changes, or other demand shaping factors in either Pearl's do-calculus \cite{PearlCausality} or Rubin's potential outcomes framework \cite{rubin1974estimating}. We should note that this is different compared to Granger causality \cite{Granger1969}, that seeks to determine the causal influence one time series has on another. In our case, price changes, promotions, and similar are interventions that we actively pursue in order to influence demand. In Pearl's do-calculus, we can express this, for example, as $P($demand $|do ($Price=2.99Euro$))$. While we can, of course, represent the prices as a chronologically ordered series of numbers, it is not really a time series but a sequence of interventions. Using an operational quantity, such as the order quantity, will in most cases act as an insufficient proxy for the quantity of interest (customer demand) and likely lead to unnecessary causal pathways that we may not be able to fully control for.

\section{Negative Binomial PDF Estimation}
\label{sec:pdfEstimation}

To predict an individual PDF using a parametric approach, one has to rely on a model assumption about the underlying distribution of the random variable to be predicted. As discussed earlier, the NBD is well routed in theoretical arguments to model customer demand. Its parameters can be modeled by two independent models, one to estimate the mean and the other for the variance. At least in principle, any method can be used. However, as discussed in sec. \ref{sec:LitRev}, machine learning algorithms are ideally suited for the task of demand forecasting and in the following, we will use the Cyclic Boosting algorithm to benefit in particular from explainable decisions rather than black-box approaches. Furthermore, the regularization approach used during the training of the Cyclic Boosting algorithm allows a dedicated treatment of the underlying NBD model, which is another major benefit compared to a standard ``off-the-shelf'' machine learning algorithm. This means we use two subsequent Cyclic Boosting models in order to estimate the parameters of each individual PDF that we need to forecast. The first model is used to estimate the mean and the second to estimate the variance. The features may or may not differ between the mean and variance estimation models, and it can be beneficial to include the corresponding mean predictions as feature in the variance model. The assigned mean and variance predictions can then be used to generate individual PDFs using the parameterization of the NBD for each sample.

In the following, after a brief recap of the fundamental ideas of Cyclic Boosting, we describe a method to predict mean and variance for individual NBDs using two Cyclic Boosting models.

\subsection{Cyclic Boosting Algorithm: Mean estimation}
\label{sec:CB}

Cyclic Boosting \cite{Wick2019} is a type of generalized additive model using a cyclic coordinate descent optimization and featuring a boosting-like update of parameters. Major benefits of Cyclic Boosting are its accuracy, performance, even at large scale, and providing fully explainable predictions, which are of vital importance in practical applications.

The main idea of this algorithm is the following: First, each feature, denoted by index $j$, is discretized appropriately into $k$ bins to reflect the specific behavior of the feature. The global mean $c$ is determined from all values $y$ of the target variable $Y \in [0,\infty)$ observed in the data. Single data records, for example, the sales corresponding to a specific product-location-date combination (meaning the sales record of a specific item sold on a specific day at a specific sales location) along with all relevant features, are indexed by $i$.
The individual predictions for the NBD mean, denoted by $\hat{\mu_i}$, can then be calculated as:
\begin{equation} \label{eqn:cb}
\hat{\mu}_i = c \cdot \prod \limits_{j=1}^p f^k_j \quad \text{with}\; k=\{ x_{j,i} \in b^k_j\}
\end{equation}
The factors $f^k_j$ are the model parameters that are determined iteratively from the features until the algorithm converges. During training, regularization techniques are applied to avoid overfitting and improve the generalization ability of the algorithm. The deviation of each factor from $f^k_j=1$ can then be used to explain how a specific feature contributes to each individual prediction.

In detail, the following meta-algorithm describes how the model parameters $f^k_j$ are obtained from the training data:
\begin{enumerate}
\item{Calculate the global average $c$ from all observed $y$ across all bins $k$ and features $j$.}
\item{Initialize the factors $f^k_j \leftarrow 1$}
\item{Cyclically iterate through features $j = 1,...,p $ and calculate in turn for each bin $k$ the partial factors $g$ and corresponding aggregated factors $f$, where indices $t$ (current iteration) and $\tau$ (current or preceding iteration) refer to iterations of full feature cycles as the training of the algorithm progresses:
\begin{equation} \label{eqn:factors}
g^k_{j,t} = \frac{\sum \limits_{x_{j,i} \in b^k_j} y_i}{\sum \limits_{x_{j,i} \in b^k_j} \hat{\mu}_{i,\tau}}\;\; \mathrm{where} \; \; f^k_{j,t} = \prod \limits_{s=1}^t g^k_{j,s}
\end{equation}
Here,  $g$ is a factor that is multiplied to the corresponding $f_{t-1}$ in each iteration. The current prediction, $\hat{\mu}_\tau$, is calculated according to \eqn \eqref{eqn:cb} with the current values of the aggregated factors $f$:
\begin{equation}
\hat{\mu}_{i,\tau} = c \cdot \prod \limits_{j=1}^p f^k_{j,\tau}
\end{equation}
To be precise, the determination of $g^k_{j,t}$ for a specific feature $j$ uses $f^k_{j,t-1}$ in the calculation of $\hat{\mu}$. For the factors of all other features, the newest available values are used, i.e., depending on the sequence of features in the algorithm, either from the current ($\tau=t$) or the preceding iteration ($\tau=t-1$).}
\item{Quit when stopping criteria, e.g., the maximum number of iterations or no further improvement of an error metric such as the MAD or MSE, are met at the end of a full feature cycle.}
\end{enumerate}

\subsection{Cyclic Boosting Algorithm: Width estimation}
\label{sec:cb_width}

In the previous section, the general Cyclic Boosting algorithm was used to estimate the mean of the NBD model. In order to predict the variance of the NBD model (associated with the mean predicted before), we modify the algorithm as follows: When looking at the demand of individual product-location-date combinations, the target variable $y$ has the values $y = 0, 1, 2, ...$ and the NBD model can be parameterized as in \cite{hilbe2011negative}:
\begin{equation} \label{eqn:nbinom}
\mathrm{NBD}(y; \mu, r) = \frac{\Gamma(r + y)}{y! \cdot \Gamma(r)} \cdot \left(\frac{r}{r + \mu}\right)^r \cdot \left(\frac{\mu}{r + \mu}\right)^y,
\end{equation}
where $\mu$ is the mean of the distribution and $r$ a dispersion parameter.

By bounding the inverse of the dispersion parameter $1/r$ to the interval $[0, 1]$ (corresponding to bounding $r$ to the interval $[1, \infty]$), the variance $\sigma^2$ can be calculated from $\mu$ and $r$ via:
\begin{equation} \label{eqn:variance_r}
\sigma^2 = \mu + \frac{\mu^2}{r}
\end{equation}

The estimate of the dispersion parameter $\hat{r}$ can then be calculated by minimizing the loss function defined in \eqn \eqref{eqn:loss_likelihood}, which is expressed as negative log-likelihood function of an NBD. The minimization over all input samples $i$ is performed with respect to the Cyclic Boosting parameters $f^k_j$, constituting the model of $\hat{r_i}$, according to \eqn \eqref{eqn:r}, where the estimates for the mean $\hat{\mu_i}$ are fixed to the values obtained from the mean model described in sec. \ref{sec:CB}.

\begin{equation} \label{eqn:loss_likelihood}
L(r) = -\mathcal{L}(r) = -\ln \sum_i \mathrm{NBD}(y_i; \hat{\mu_i}, \hat{r_i})
\end{equation}

\begin{equation} \label{eqn:r}
\hat{r_i} = 1 +  \frac{1}{\prod \limits_{j=1}^p f^k_j} \quad \text{with}\; k=\{ x_{j,i} \in b^k_j\}
\end{equation}

In other words, the values $\hat{r_i}$ are estimated via learning the Cyclic Boosting model parameters $f^k_j$ for each feature $j$ and bin $k$ from data. For any concrete observation $i$, the index $k$ of the bin is determined by the value of the feature $x_{j,i}$ and the subsequent look-up into which bin this observation falls. Like in sec. \ref{sec:CB}, the model parameters $f^k_j$ correspond to factors with values in $[0, \infty]$, and again values deviating from $f^k_j=1$ can be used to explain the relative importance of a specific feature contributing to individual predictions. Note that the structure of \eqn \eqref{eqn:r} can be interpreted as inverse of a logit link function in the same way as explained in \cite{Wick2019} when Cyclic Boosting is used for classification tasks.

In the same way as described in sec. \ref{sec:CB} for its basic multiplicative regression mode, the Cyclic Boosting algorithm is trained iteratively using cyclic coordinate descent, processing one feature with all its bins at a time until convergence is reached. However, unlike in \eqn \eqref{eqn:factors} of the basic multiplicative regression mode, the minimization of the loss function in \eqn \eqref{eqn:loss_likelihood} cannot be solved analytically and has to be done numerically, for example, using a random search. All other advantages of Cyclic Boosting, like individual explainability of predictions, remain valid for its negative binomial width mode.

Finally, the variance $\hat{\sigma}^2_i$ can be estimated from the dispersion parameter $\hat{r_i}$ using \eqn \eqref{eqn:variance_r}. And together with the individual predicted mean $\hat{\mu_i}$ from the first step, the NBD model is then fully specified for each individual prediction $i$.

\section{Evaluation of PDF Predictions}
\label{sec:pdfEvaluation}

Many statistical and most machine learning methods do not provide a full PDF as result. Instead, these methods typically predict a single numerical quantity (usually denoted by $\hat{y}$, in this work denoted by $\hat{\mu}$ to make evident that it is the prediction of the mean of an underlying probability distribution) that is then compared to the observed concrete realization of the random variable (denoted by $y$) using metrics such as the MSE, MAD, or others. In the setting of a retailer, the observed quantity is the sales of individual products and most approaches would then predict a single number as a direct estimate of the sales. However, reducing the prediction to a single number does not allow to account for the uncertainty of the prediction or the dynamics of the system. Instead, it is imperative to predict the full PDF for each prediction to be able to optimize the subsequent operational decision. Unfortunately, most statistical or machine learning methods that predict full individual probability functions lack quantitative or even qualitative evaluation methods to assess whether the full distribution has been forecasted correctly, in particular in its tails.

In the simplest case, we only have one model with one set of model parameters to cover all predictions. In this case, the evaluation of the full PDF is straightforward: We would fill a histogram of all observed values, such as sales records, and overlay this with the single model, such as an NBD with predicted parameters, that is used for all observations. Then, we compare the model curve directly with the observations, using statistical tests such as the Kolmogorov-Smirnov test.

In practical applications however, we have a large number of effective prediction models, since although we may always use the same model parameterization, such as the NBD, its parameters have to be determined at the required level of granularity. For example, for daily orders, we need to predict the parameters of the NBD for each product, location, and sales day. Unlike the simple case discussed above, where we had many observations to compare the prediction model to, we now have just a single observation per prediction, meaning that we cannot use statistical tests directly.

For an estimation of the determining parameters of an assumed functional form for the PDF, assessing the correctness of the PDF model output refers to the evaluation of the accuracy of the prediction of the different parameters. In the case of the NBD used in this work, we have to verify that mean and variance are determined accurately, as well as checking that the choice of the underlying model can describe the observed data.

In the following, we will show how different visualizations of the observed cumulative distribution function (CDF) values can be used to evaluate the quality of the predicted PDFs. Although we limit the following discussion to the negative binomial model, the method can be applied generally to any representation of a PDF, even if it is obtained empirically.

\subsection{Histogram of CDF Observations}
\label{sec:cdf_histo}

For a first qualitative assessment, we make use of the probability integral transform, see, e.g., \cite{Angus1994,casella2002statistical}, which states that a random variable distributed according to the CDF of another random variable is uniformly distributed between 0 and 1. We therefore expect that the distribution of the actually observed CDF values of the corresponding individual PDF predictions is uniform, if the predicted PDF is calibrated correctly, {\em regardless} of the shape of the predicted distribution. Any deviation can be interpreted as a hint that the predicted PDF is not fully correct \cite{diebold1998vevaluating}.

\noindent
The CDF of a PDF $f(x)$ is defined as:

\begin{equation}
\label{eqn:CDF}
F_X(x) = P(X \le x) = \int_{-\infty}^{x} f_X(x^\prime) dx^\prime
\end{equation}

Here, $F_X(x)$ is the CDF with $\lim_{x \to -\infty}F_X(x) = 0$ and $\lim_{x \to \infty}F_x(x) = 1$. The cumulative distribution describes the probability that the variable has a value smaller than $x$, and intuitively represents the area under $f(x^\prime)$ up to a point $x$.

If the CDF is continuous and strictly increasing, then the inverse of the CDF, $F^{-1}(y)$, exists and is a unique real-valued number $x$ for each $y \in [0,1]$, so that we can write $F(x) = y$. The inverse of the CDF is also called the quantile function, because we can define the quantile $q$ of the PDF $f(x)$ as:

\begin{equation}
Q_q = F^{-1}(q)
\end{equation}

\begin{figure}
\begin{center}
\includegraphics[width=8cm]{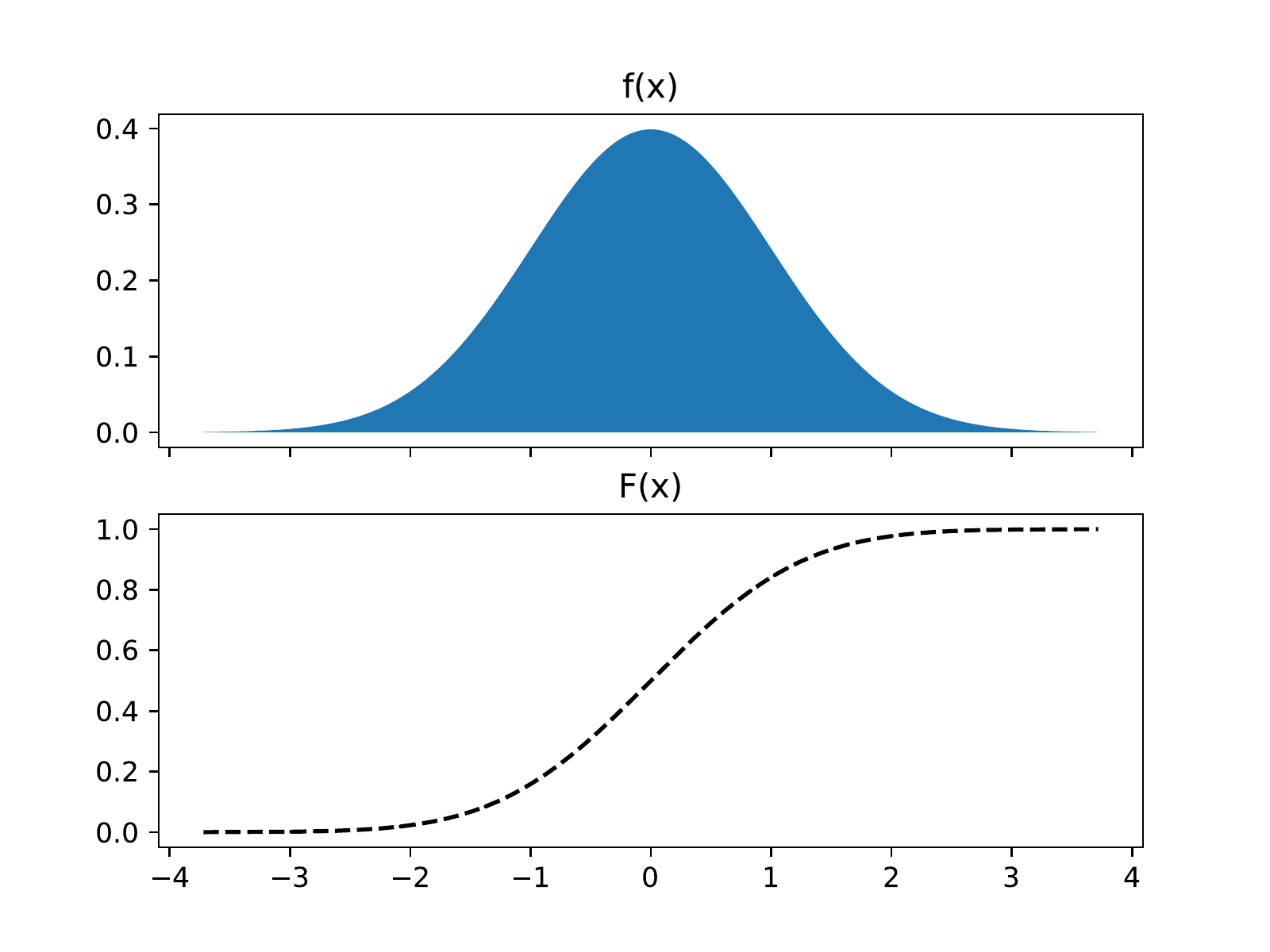}
\caption{\label{fig:PdfCdf} PDF $f(x)$ and CDF $F(x)$ of a normal distribution.}
\end{center}
\end{figure}

Using the example of the normal distribution with $\mathcal{N}(0,1)$ as shown in \fig \ref{fig:PdfCdf}, we can identify the median ($q = 0.5$) by first looking at the CDF in the lower part of the figure, look at the value $0.5$ on the $y$ axis and then identify the point on the $x$ axis for both the PDF $f(x)$ and the CDF $F(x)$ that corresponds to the quantile $q$. In the case of the normal distribution, this is of course the central value at zero.

We can then interpret the CDF as a new variable $s = F(t)$, meaning that $F$ becomes a transformation that maps $t$ to $s$, i.e., $F:t \to s$. Accordingly,  $\lim_{t \to -\infty}s(t) = 0$ and $\lim_{t \to \infty}s(t) = 1$, and $s$ can be intuitively interpreted as the fraction of the distribution of $t$ with values smaller than $t$ from the definition of the CDF. This implies that the PDF of $s$, $g(s)$, is constant in the interval $s \in [0,1]$ in which $s$ is defined, and $s$ can be interpreted as the CDF of its own PDF:

\begin{equation}
s = G(s) = \int_{-\infty}^{s} g(s^\prime) ds^\prime
\end{equation}

In case of discrete probability distributions, such as the NBD, the same argument still holds, but the definition of the quantile function is replaced by the generalized inverse: $F^{-1}(y) = \mathrm{\inf \left \{x : F(x)>y\right  \} }$ for $y \in [0,1]$, see e.g., \cite[p. 54]{casella2002statistical}. In order to obtain a uniform distribution for discrete PDFs that is comparable to the case of continuous distributions, the histogram holding the values of the CDF is filled using random numbers according to the intervals of the CDF. For example, if the sales of zero items account for 75 percent of the observed sales distribution for this item, the value of the CDF function that is used to fill the histogram in case of zero actual sales is randomly chosen in the interval $[0, 0.75]$. Proceeding similarly for all other observed values, with the intervals from which to randomly choose values to fill in the histogram defined by the CDF values of the corresponding discrete sales value and the one below (e.g., for 3 actual sales: random pick between discrete CDF values for 2 and 3), the resulting histogram of CDF values is again uniform, as in the case of a continuous PDF.

\begin{figure}
\begin{center}
\includegraphics[scale=0.25]{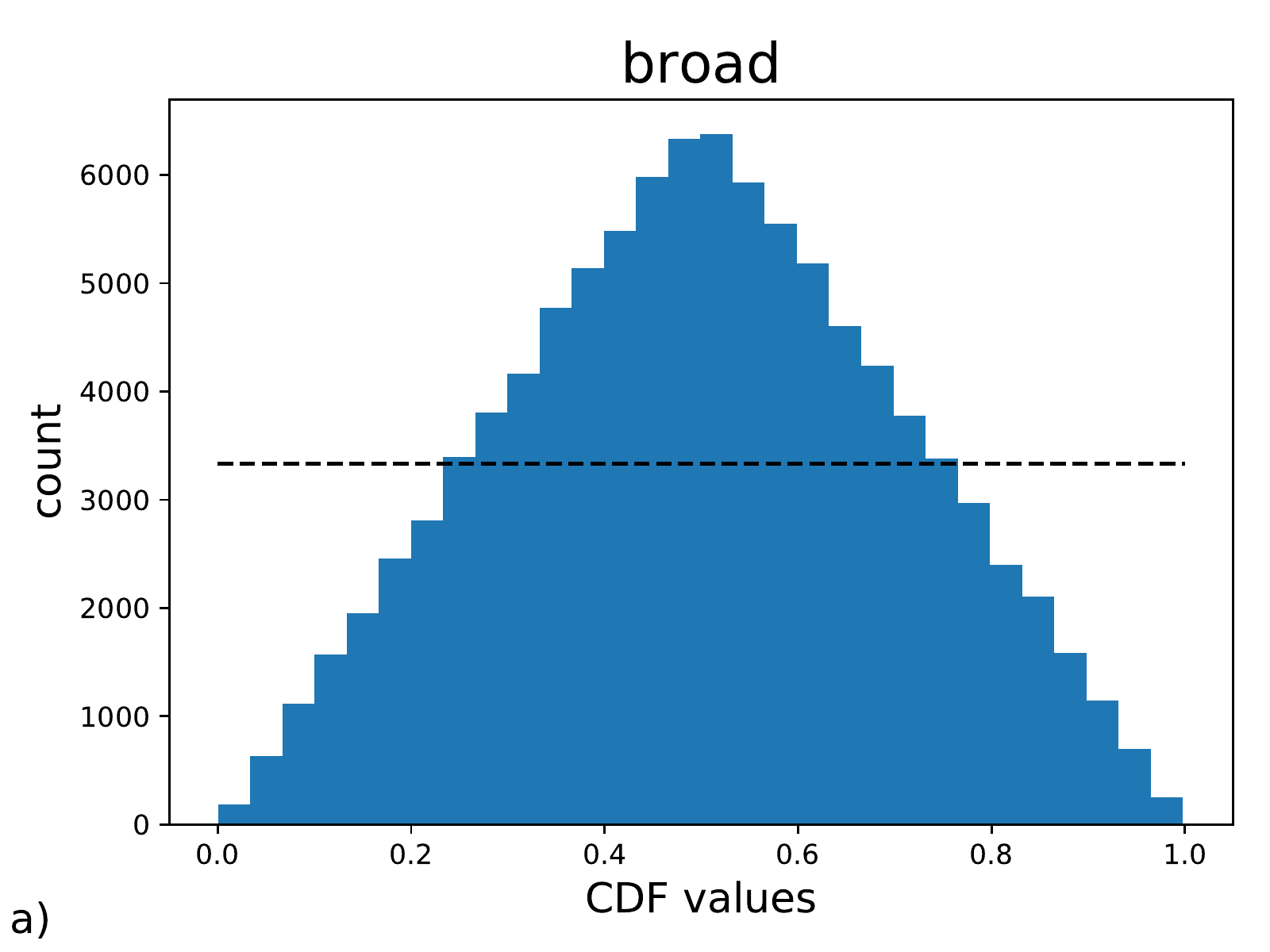}
\includegraphics[scale=0.25]{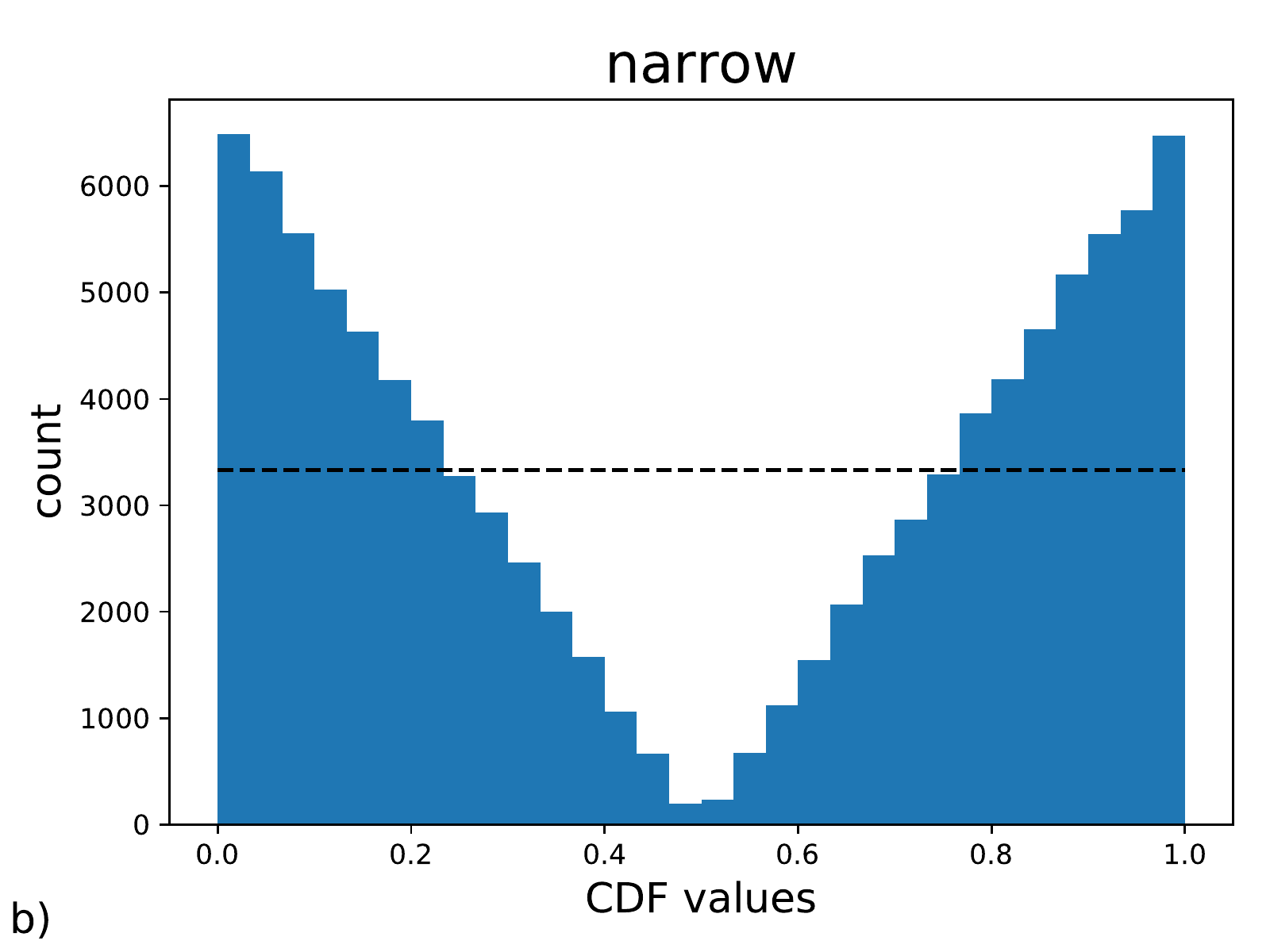}
\includegraphics[scale=0.25]{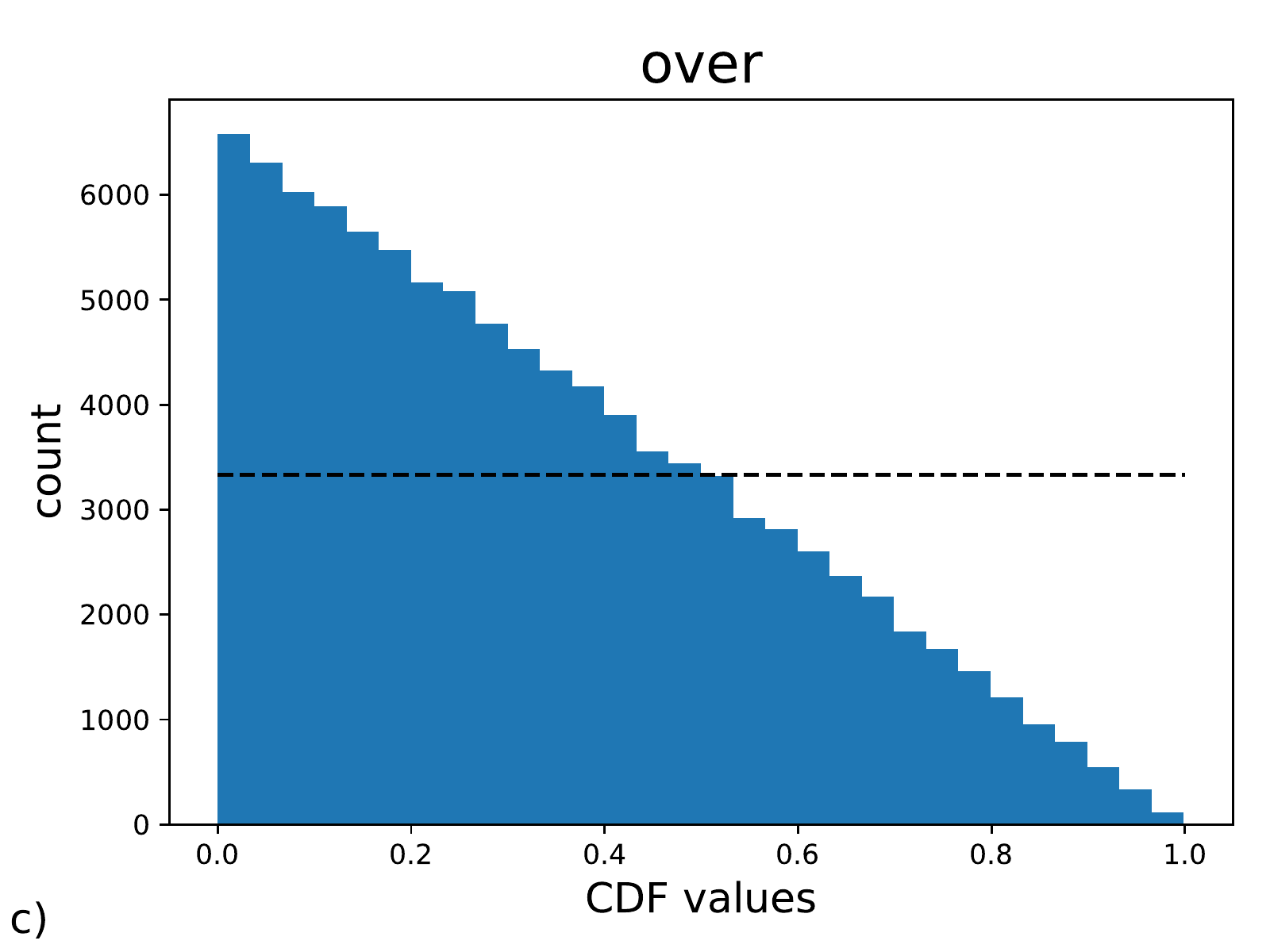}
\includegraphics[scale=0.25]{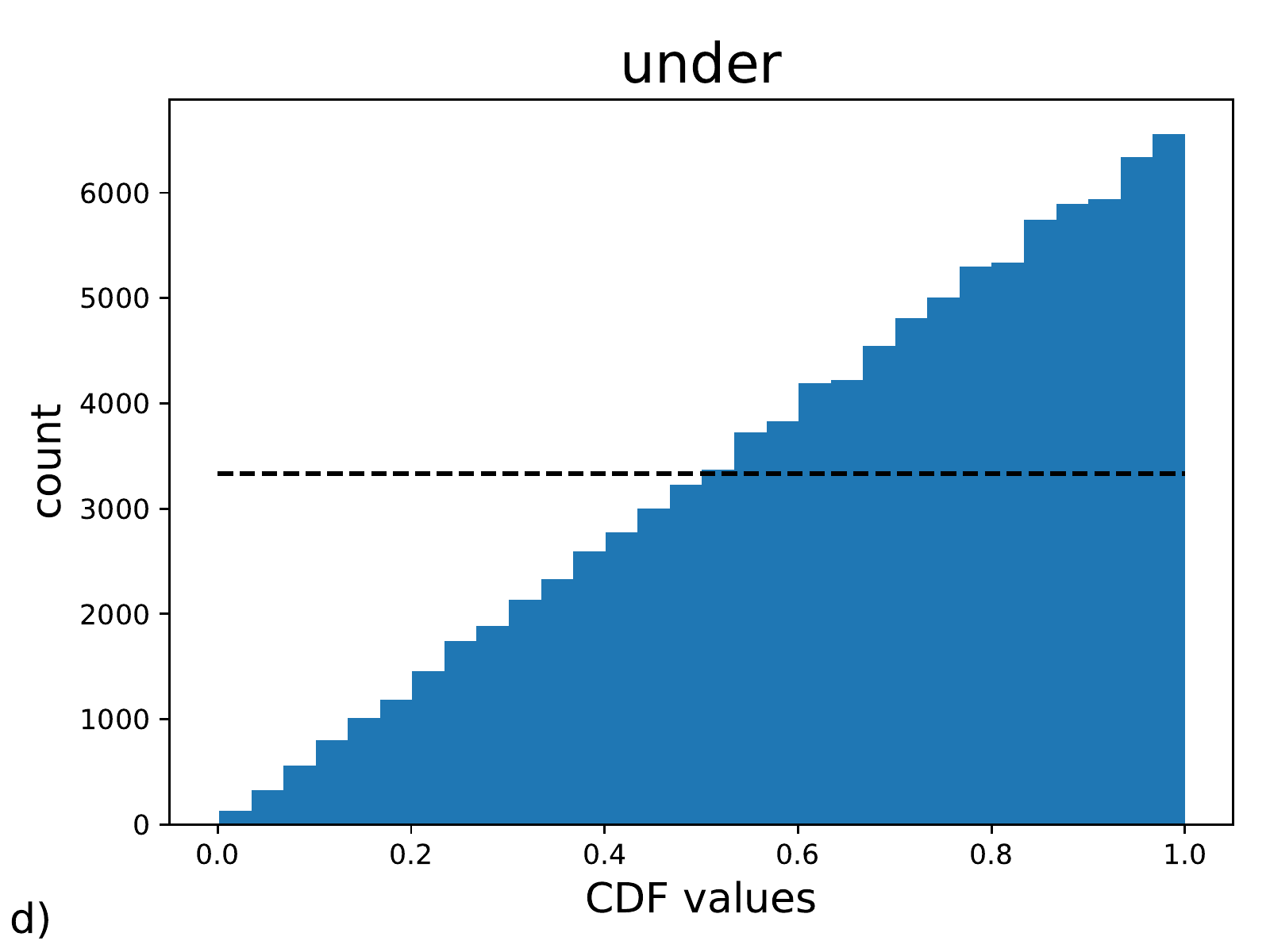}
\caption{\label{fig:cdf_histos} Histogram of actually observed CDF values for different cases of estimating the full PDF compared to the expected uniform distribution (dashed horizontal line in each of the plots): variance overestimated (a: ``broad''), variance underestimated (b: ``narrow''), mean overestimated (c: ``over''), mean underestimated (d: ``under'')}
\end{center}
\end{figure}

A histogram of the actually observed CDF values for each individual PDF prediction (see \fig \ref{fig:pdf_example} for an example) is therefore expected to be uniformly distributed in $[0,1]$, if the predicted PDF $f(x)$ is correctly calibrated, i.e., if both the choice of the model and the model parameters are estimated correctly. If the mean or the variance are not estimated correctly, the resulting distribution will show a distinct deviation from this uniform behavior. This is illustrated in \fig \ref{fig:cdf_histos}, which shows the distribution of observed CDF values for four different cases.

It should be noted that the method requires a sufficient sample size, as a too low number of observations of course leads to a discretization bias.

\subsection{Inverse Quantile Profile Plot}
\label{sec:invquant_plot}

We now refine the method from sec. \ref{sec:cdf_histo} by comparing (in the sense of higher or lower) the CDF values of the observed events, i.e., the sales records, with specified quantiles. In order to do so, we start again from the predicted values for the mean and variance from which an NBD is constructed for each prediction, for example, for the predicted demand for a specific product sold on a single day in a given sales location. Each of these predicted negative binomial PDFs is then transformed to its CDF. Note that for simplicity, we always refer to the negative binomial model in this description, however, the general approach is valid for any PDF.

We then compare the actual observed sales value (corrected for censored data if necessary) to different quantiles of the corresponding predicted distribution for each data record and average over a larger data sample. For example, if we wanted to check that the median of the distribution, corresponding to the quantile $0.5$, is predicted correctly by the machine learning model, we would compare the value $0.5$ to the ratio of CDF values (again randomly chosen from the corresponding range of CDF values for discrete target values, as described in sec. \ref{sec:cdf_histo}) of observed sales records being lower/higher than $0.5$. In other words, in case of the median, 50 percent of the ex post observed target values should be observed below the median of the corresponding individually predicted PDF and 50 percent above.

In order to judge whether the overall shape of the predicted distributions is estimated correctly, we repeat this procedure for a range of quantiles, for example, $q = 0.1, 0.2, \ldots, 0.9$. However, we are free to choose which quantiles to look at, and in specific situations it might be advisable to look at the tails of the distribution in more detail, to make sure that even relatively rare events are estimated correctly by the machine learning algorithm, and add more quantiles for comparison in the region between, say, $q = 0.95$ and $q = 0.99$. In the following, we call this method {\em inverse quantile profile plot}. Profile plots are akin to scatter plots and described in more detail in Appendix \ref{sec:profile}.

\fig \ref{fig:invquant_example} illustrates five different collections of inverse quantile profile plots (each collection comparing to 6 specified quantile values, namely $q = 0.1$, $q = 0.3$, $q = 0.5$, $q = 0.7$, $q = 0.9$, and $q = 0.97$), for separate sets of exemplary PDF estimations and observed data combinations. The dashed horizontal lines indicate the fraction we expect, i.e., the specified quantile value, if the predictions are correct. For example, for the median, the line at $q = 0.5$ indicates that  50 percent of all PDF prediction and observed data combinations in a given data set should fall above the line, and 50 percent should fall below the line. The observation of the number of samples, indicated with different markers, that do in fact fall above and below a particular line, then allows the evaluation of the accuracy of PDF estimations. In case that the PDFs are not estimated correctly, the fractions will deviate from their expected values and the corresponding profile plot allows to judge whether for example the tails of the predicted distribution describing rare events are particularly problematic.

\begin{figure}
\begin{center}
\includegraphics[width=8cm]{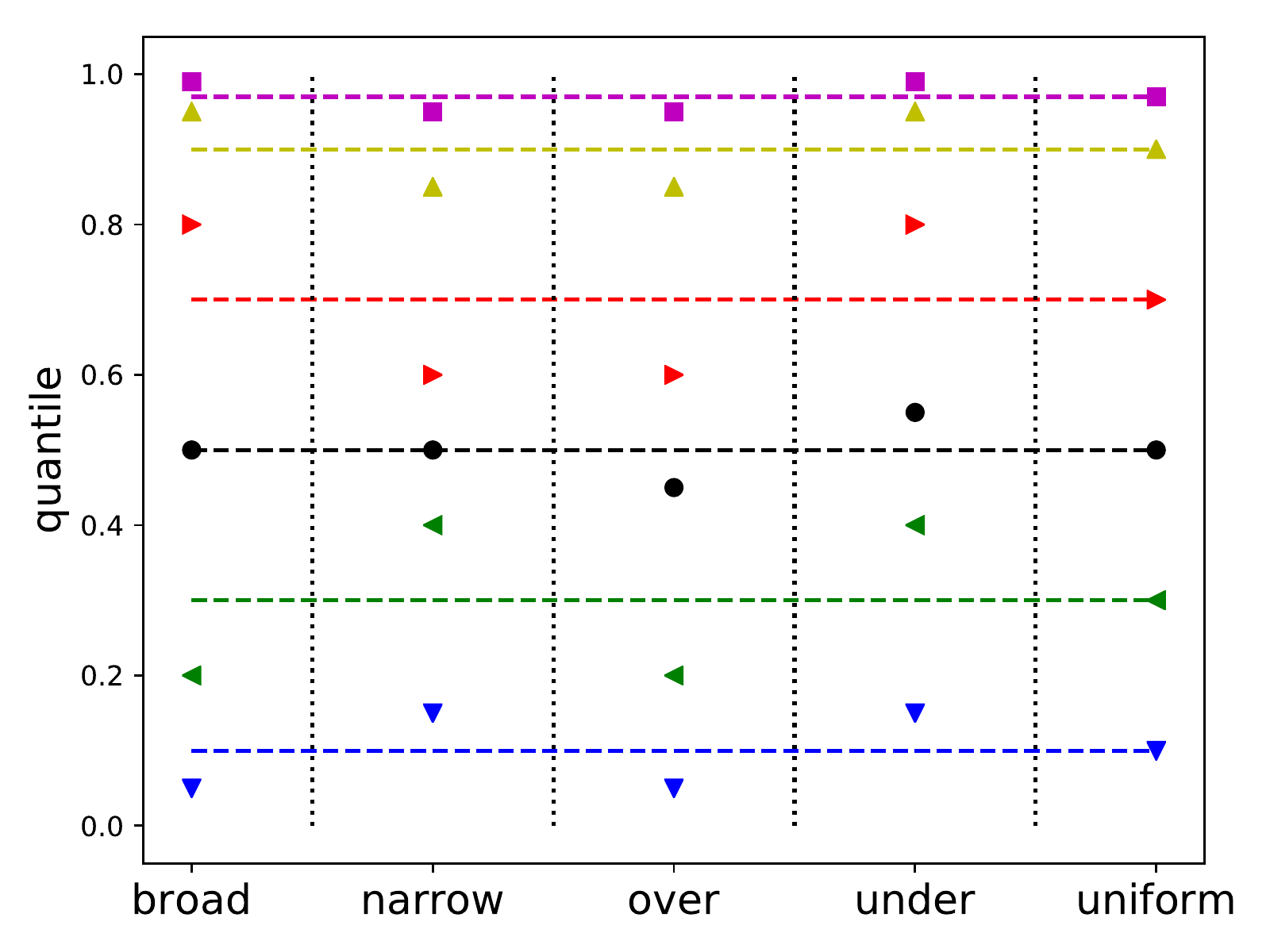}
\caption{\label{fig:invquant_example} Five different inverse quantile profile plots (separated into columns by dotted vertical lines), each comparing all its predicted PDFs to the corresponding observed events. In the leftmost two columns, the outcome is illustrated if the estimate of the variance is biased, the center and center-right columns illustrate cases for which the mean estimation is biased, and the far right column shows the expected behavior if all predictions are correct.}
\end{center}
\end{figure}

In the same way as the method of filling all observed CDF values of the individual PDF predictions in a histogram and comparing to a uniform distribution (described in sec. \ref{sec:cdf_histo}), inverse quantile profile plots do not work on individual PDF predictions but require a statistical population. However, calculating the fractions of the inverse quantile profile plots globally, i.e., over all samples in the data set, might not reveal certain deficiencies for a subset, e.g., a specific store or product group in the example of retail demand forecasts. Therefore, we combine the approach described above with the method of profile plots, where the quantity on the $x$ axis of the inverse quantile profile plot can be any variable of the data set at hand known at prediction time.

In summary, this approach has two major benefits compared to the method discussed in sec. \ref{sec:cdf_histo}: First, by explicitly visualizing several different quantiles, the inverse quantile profile plot reveals which part of the predicted PDF, for example the tails, are particularly problematic. Second, by showing the dependency from the (arbitrary) variable on the $x$ axis, the inverse quantile profile plot reveals deviations of the predicted PDF from the actuals for different parts, e.g., specific categories, of that variable. Two examples for this can be found in Figs. \ref{fig:invquant_dayofweek} and \ref{fig:invquant_mean} in the next section.

\subsection{Quantitative Evaluation of PDF Predictions}
\label{sec:cdf_acc}

The methods discussed so far allow a detailed qualitative evaluation of PDF predictions. In order to also quantify the quality of the PDF predictions, we use a metric assessing the difference between two probability distributions to compare the histogram of CDF observations of the predicted PDFs with the expected uniform distribution, and define an accuracy measure in the range between 0 and 1, which takes the value of 1 when both distributions agree perfectly. Several approaches that measure the difference between two probability distributions are suggested in the literature, such as the first Wasserstein distance \cite{olkin1982}, also known as earth mover distance (EMD), the Kullback-Leibler divergence \cite{kullback1951}, also known as relative entropy, the Jensen-Shannon divergence \cite{dagan1997}, also known as information radius, or the energy distance \cite{SZEKELY20131249}.

Compared to the Kullback-Leibler and Jensen-Shannon divergences, the first Wasserstein distance, representing the symmetric distance between two PDFs on a given metric space, is more sensitive to smaller deviations, because it exhibits a linear behavior around zero (reflecting perfect agreement). Therefore, we focus on the first Wasserstein distance as the measure of difference in the following. For our purposes here, it can be defined by:
\begin{equation}
\text{EMD}(P, Q) = \frac{\sum_{k=1}^N |F_P(x_k) - F_Q(x_k)|}{N},
\end{equation}
where $F_P(X)$ and $F_Q(X)$ are the CDFs of the two PDFs $P(X)$ and $Q(X)$, respectively, and $x_k$ denotes the average value of $X$ in bin $k$, with $X$ being divided in $N$ bins.

Since $0.5$ represents the maximum value of the first Wasserstein distance when comparing any distribution in the support $[0, 1]$ to a flat distribution in the same interval (its minimum being zero), we define an accuracy measure for our PDF predictions in the range $[0, 1]$ by:
\begin{equation}
\text{accuracy} = 1 - 2 \cdot \text{EMD}
\end{equation}

\section{Temporal Confounding in Time Series Forecasting}
\label{sec:ts_confounding}

\subsection{Origin of Temporal Confounding}
\label{sec:IntroTempConfound}

The aim of time series forecasting in general is to predict the next values in a series of time-ordered events. The quantity we wish to predict is often called the ``prediction target'' or just ``target''. In the case of demand forecasting, the prediction target is the expected future demand, and we usually employ the time series of observed sales events in the past, possibly corrected for out-of-stock situations, to estimate the model parameters. Many time series forecasting methods exploit the autocorrelation between the values of the variable of interest at different times. Depending on the exact nature of the problem we consider in these approaches, different lags between the variable $Y$, describing the demand, play an important role. For example, a weekly repeating pattern would lead to a strong autocorrelation for a lag of $7$ (days), meaning that the values $y_t$ and $y_{t-7}$ of the variable $Y$ would be strongly autocorrelated. However, this autocorrelation between different times of the target variable may not be a true causal dependency, but a spurious correlation due to one or more temporal confounders. For example, if an exogenous variable has a direct causal impact on the variable $Y$ at time, say, $t-1$, this external variable will often also have an impact at time $t$. Intuitively, if the variable $Y$ was influenced yesterday by some external event, there is a high chance that we can observe this impact today as well.

\begin{figure}
\begin{center}
\includegraphics[width=5cm]{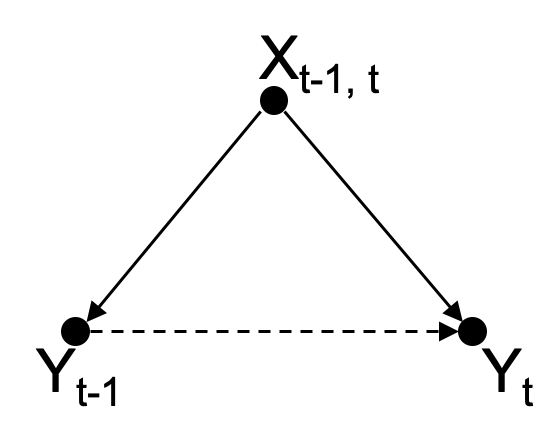}
\caption{\label{fig:temporal_confounding} Temporal confounding by an exogenous variable $X_{t-1, t}$ stable over the time steps $t-1$ and $t$, creating a spurious autocorrelation between $Y_{t-1}$ and $Y_t$.}
\end{center}
\end{figure}

The effect of temporal confounding is illustrated in \fig \ref{fig:temporal_confounding}, where the variable $Y$ is observed at two different times $t-1$ and $t$ with values $y_{t-1}$ and $y_t$, and the variable $X$ is an external influencing factor that affects the variable $Y$ both at time $t-1$ and at time $t$. For example, we can imagine that $X$ represents the price of a product that is changed due to a special promotion at these two time steps. Looking at the causal structure of the directed acyclic graph (DAG) created by this dependency, we see that  $X$ takes the role of a (temporal) confounder that creates a backdoor path between $Y_{t-1}$ and $Y_t$. This in turn means that there is now a spurious (auto)correlation between $Y_{t-1}$ and $Y_t$, indicated by the dashed line in \fig \ref{fig:temporal_confounding}. Therefore, if we do not control for the temporal confounder $X$, but use the autocorrelation between $y_{t-1}$ and $y_t$ to forecast the next value $y_{t+1}$, we rely on a spurious correlation between $Y_{t-1}$ and $Y_t$ that is not ``real'' in the sense that it results from a causal influence of $Y_{t-1}$ on $Y_t$, but originates from the backdoor path via $X_{t-1, t}$.

In the case of retail demand forecasting, we typically have many temporal confounders, as demand is influenced by a multitude of different causes such as prices, advertisements, weather, and many more. Some of these are due to interventions, whereas others are caused by external effects we cannot influence. For example, setting the price of a product or running an advertisement campaign is an intervention (in the sense of causal analysis) by the retailer, whereas the impact of the forecasted weather at the sales location and sales time on the expected demand is outside our control. And in addition to the spurious autocorrelations due to all these temporal confounders, there might also be a direct impact of $Y_{t-1}$ on $Y_t$, resulting in an autocorrelation originating from a ``real'' cause. If we can control for the confounder(s) $X$, we can remove, or at least limit, the spurious autocorrelation and in turn improve the forecasting method, especially in terms of its generalizability, by focusing on the description of the remaining ``real'' autocorrelation.

\subsection{Separation of Causal Features from Autocorrelation in Machine Learning Models}

% time series with ML and ewma features
Using machine learning, we express the task of time series forecasting as a supervised learning approach, where the prediction target, i.e., the quantity we wish to forecast, is represented by the values of the time series of interest itself. In the case of demand forecasting, the future expected demand is the variable we wish to forecast, and we need to build feature variables, either by hand or using some automatic approach, to construct a supervised machine learning model. A common approach is to exploit the autocorrelation of the target time series variable to build dedicated feature time series from its summary statistics of the respective recent past. For example, we could use an exponentially weighted moving average (EWMA), calculated according to \eqn \eqref{eqn:ewma}, for each individual time series and choose several different values for the smoothing parameter $\alpha$, which is defined in the range $[0, 1]$.
\begin{equation} \label{eqn:ewma}
\begin{split}
&\mathrm{EWMA}(x_{t+\mathrm{horizon}}) = \\ &\frac{x_t + (1 - \alpha) x_{t-1} + (1 - \alpha)^2 x_{t-2} + ... + (1 - \alpha)^t x_0}{1 + (1 - \alpha) + (1 - \alpha)^2 + ... + (1 - \alpha)^t}
\end{split}
\end{equation}
We can then use these EWMA features in the supervised machine learning model that we use to predict the future value of the target variable, in our case, the future demand for a given product at a given location. When constructing these features, we need to keep in mind that we use a lag with respect to the target time series corresponding to the forecast horizon. For example, if we were to forecast three days ahead into the future, the lag of the feature time series with respect to the target time series has to be three as well. By creating many variants of the lagged target information contained in the original time series, for example by varying the value of the hyperparameter $\alpha$ or calculating distinct moving averages for the different days of the week, we can build a wide range of variables capturing the dynamics of the recent past and use these as features for the machine learning model. In addition, we can include further covariates in the model, such as pricing information, promotion details, characteristics of the products we wish to forecast, or characteristics of the location at which the product is sold.

% problem statement
Although this approach is often taken in practice, it is prone to a detrimental effect of temporal confounding: Because we do not control for temporal confounders and exploit the autocorrelation of the target time series using its lagged information directly in form of, for example, an EWMA, we obscure the underlying causal dependencies and leave the temporal backdoor paths created by them open. This in turn makes learning the true causal dependencies between the exogenous variables and the quantity of interest much more difficult, because the machine learning algorithm can now learn the statistical dependencies of the spurious autocorrelation encoded in the time series features, rather than learn the more fundamental influences from variables such as price information, weather conditions, and similar. For example, if we wish that the model learns the dependency, say, the price of a product or a promotion has on the customer demand for the product, we provide this information twice: Once by explicitly adding the price or the promotion details as features to the machine learning model, and then again implicitly by using lagged target information, such as EWMA features of the observed sales in the past. Looking back at \fig \ref{fig:temporal_confounding}, the EWMA contains the spurious autocorrelation created by the backdoor path via the exogenous variable(s) (such as the price or promotion details). The machine learning algorithm is not aware of the causal structure and its influences on the time series, but only ``sees'' the correlation to the prediction variable. Therefore, the machine learning algorithm is prone to learn the spurious autocorrelation encoded in the EMWA features, as it cannot distinguish these from the true causal influences.

As traditional statistical forecasting or machine learning methods generally rely upon statistical dependencies, no matter if spurious or from true causal effects, this does not seem to be a problem for the predictive power at first glance. However, it can have severe impacts on the explainability and generalizability of the model, because the autocorrelation, being a composed effect of many causes, is less stable over time than the actual causal effects. A concrete weakness of the reliance on autocorrelation can be seen when looking at effects at isolated time steps. For example, if the target variable has a sharp peak that is induced by a sudden change of an exogenous variable with a causal influence on the target, the autocorrelation, lagged by its nature, would not help to predict the peak at the very day it occurs, but would tend to ``trick'' the model in predicting a delayed peak, which in turn reduces the accuracy of the prediction at other time steps after the actual peak as well.

Note that the effect of temporal confounding does not only apply to this specific approach of building features such as the EMWA to be used in supervised machine learning, but to all time series forecasting methods that rely on the autocorrelation between events at various lags, including ARIMA, Holt-Winters, Kalman filters and the related variants, RNNs and its variants like LSTMs, and others summarized in sec. \ref{sec:LitRev}.

% separate
In order to avoid these issues, we suggest to separate the target autocorrelation from the exogenous dependencies. This can be done by avoiding any features in the machine learning model that include lagged target information. Using the example from above, this would mean that we include features such as the price of the product or promotion details, weather information, as well as other variables describing the characteristics of a given product at a specific sales location, but do not include lagged information from the sales time series, such as its EWMA or similar approaches.

Therefore, the features do not include any variable that is directly constructed from the previously observed sales events. As a consequence, the machine learning model needs to learn from the true causal dependencies expressed in the exogenous features, which improves both the explainability of the predictions, as well as the generalizability of the prediction model, because the causal dependencies are not confused by additional spurious autocorrelations. It should be noted that variables describing certain seasonal effects, like the day of the week or the day of the year, can be included in the machine learning model without any negative effects. In fact, these represent seasonal causal dependencies of the target.

We need to keep in mind that, although this approach of handling temporal confounders improves the prediction model used for demand forecasting, it is not yet sufficient for true causal inferences. For this, we would also have to control for the confounding effects between the different exogenous features of the machine learning model and its prediction target.

\subsection{Autocorrelated Residual Correction}

By including all available exogenous variables as features in the machine learning model but excluding any endogenous information from the target autocorrelation we can, at least partially, control for the temporal confounders. Any remaining autocorrelation between different lags of the time series of the residuals formed from the original time series and the time series of the forecasted values from the machine learning model discussed above can therefore originate from two reasons: Either there is a genuine autocorrelation between different lags $Y_{t-\mathrm{lag}}$ and $Y_t$ of the variable we wish to forecast, or we have not included all relevant causal variables in the machine learning model. To make use of this remaining autocorrelation, we suggest to apply a residual correction on each of the predictions of the machine learning model $\hat{\mu}_{\mathrm{ML}}$ (assuming predictions of the mean of an underlying PDF as discussed in sec. \ref{sec:pdfEstimation}), based on the deviations in the recent past between the target and these predictions. Such a residual correction heuristically captures the recent trends without modeling them explicitly, and can be seen as an empirical correction to improve the final forecast quality.

% ewma
One possibility for the implementation of the residual correction is to apply exponential smoothing. In the simplest case, this can be done as stated in eqn. \ref{eqn:emov_correction}, where the two EWMA terms are grouped by the individual time series and calculated according to eqn. \ref{eqn:ewma}.
\begin{equation} \label{eqn:emov_correction}
\hat{\mu} = \frac{\mathrm{EWMA}(y)}{\mathrm{EWMA}(\hat{\mu}_{\mathrm{ML}})} \cdot \hat{\mu}_{\mathrm{ML}}
\end{equation}
In most cases, simple exponential smoothing will be sufficient for the residual correction, since we have already incorporated all relevant external effects, including seasonalities like a dependency of the target on the day of the week, in the machine learning model before, and only seek to capture the information that we have missed, because, for example, we do not have the relevant data to describe this as a further feature variable. However, we need to be aware that if we have missed a crucial temporal confounder when building the features for the machine learning model, applying this residual correction risks opening a backdoor path for this confounder, which in turn might result in slightly sub-optimal predictions, as discussed in sec. \ref{sec:IntroTempConfound}. We should therefore take care to use the relevant domain knowledge to capture the causal structure of the problem at hand as best as we can. On the other hand, the method proposed above does allow us to take temporal confounders into account at all, and we are only limited by the extend of our domain knowledge, or to capture the relevant data, to identify and include the temporal confounders as exogenous feature variables in the machine learning model.

\section{Example: Retail Demand Forecasting}
\label{sec:example}

In the following, we describe how to use the approach outlined above in a practical setting. We use a public dataset obtained from a Kaggle online competition focusing on estimating unit sales of Walmart retail goods \cite{kaggle_data} for individual items for specific stores on specific days. For each demand forecast, Cyclic Boosting is used to predict the full probability distribution of the expected demand at a granularity of (item, store, day) and use the methods described in sec. \ref{sec:pdfEvaluation} to evaluate the quality of the individual forecasts. Each data record corresponding to an observed sales record is described by the following fields: the identifier of an individual store (\texttt{store\_id}), the product identifier (\texttt{item\_id}), and the date(\texttt{date}). The target $y$, that we need to predict, is the number of sales of a given product in a given store on a specific day, denoted by \texttt{sales}.

For our experiments, we use data from 2013-01-01 to 2016-05-22, that describe the sales of 100 different products (\texttt{FOODS\_3\_500}, ..., \texttt{FOODS\_3\_599}) of the department \texttt{FOODS\_3} in all 10 available stores.  All data before 2016 are used as the training data, and the data from 2016 are used as an independent test set. Besides the fields used to identify an individual sales record and the corresponding observed sales value, namely \texttt{item\_id}, \texttt{store\_id}, \texttt{date}, \texttt{sales}, we also use the fields \texttt{event\_name\_1}, \texttt{event\_type\_1}, \texttt{snap\_CA}, \texttt{snap\_TX}, \texttt{snap\_WI}, \texttt{sell\_price}, and \texttt{list\_price}, and multiple features built from these variables are then included in the machine learning models.

\subsection{Mean Estimation}
\label{sec:example_mean}

As discussed earlier, we assume that each individual sale can be described by a Poisson-like process and we assume an NBD to model the individual probability distribution of each sales event. As a first step, we use the supervised machine learning method Cyclic Boosting (as described in sec. \ref{sec:CB}) to predict the mean of the distribution.

This model uses the following variables as features: categorical variables for \texttt{store\_id} and \texttt{item\_id}, several derived variables that are constructed from the time series of the sales records describing trend and seasonality (days since beginning of 2013 as linear trend as well as day of week, day of year, month, and week of month), time windows around the events given in the data set (7 days before until 3 days after for Christmas and Easter, and 3 days before until 1 day after for all other events like New Year or Thanksgiving), a flag denoting a promotion, and the ratio of reduced (\texttt{sell\_price}) and normal price (\texttt{list\_price}). We also include various two-dimensional combinations of these features. In these cases, one of the two dimensions is either \texttt{store\_id} or \texttt{item\_id}, allowing the machine learning model to learn characteristics of individual locations and products.

\noindent
As discussed in sec. \ref{sec:ts_confounding}, we do not use any lagged target information, for example via the inclusion of target EWMA features, in the machine learning model, but apply an individual residual correction on each of its predictions, which accounts for deviations between the EWMA (with $\alpha=0.15$) of the predictions and targets of each product-location combination over the corresponding past. Empirically, a value of $\alpha=0.15$ gives good results here, though in a practical application, this parameter would have to be optimized using methods such as cross-validation according to a metric that reflects the objective of the business strategy. To reflect a realistic replenishment scenario, we use the model to predict the expected demand two days into the future, meaning that we use a lag of two days for the calculation of both the target EWMA and the prediction EWMA.

\begin{figure}
\begin{center}
\includegraphics[width=8cm]{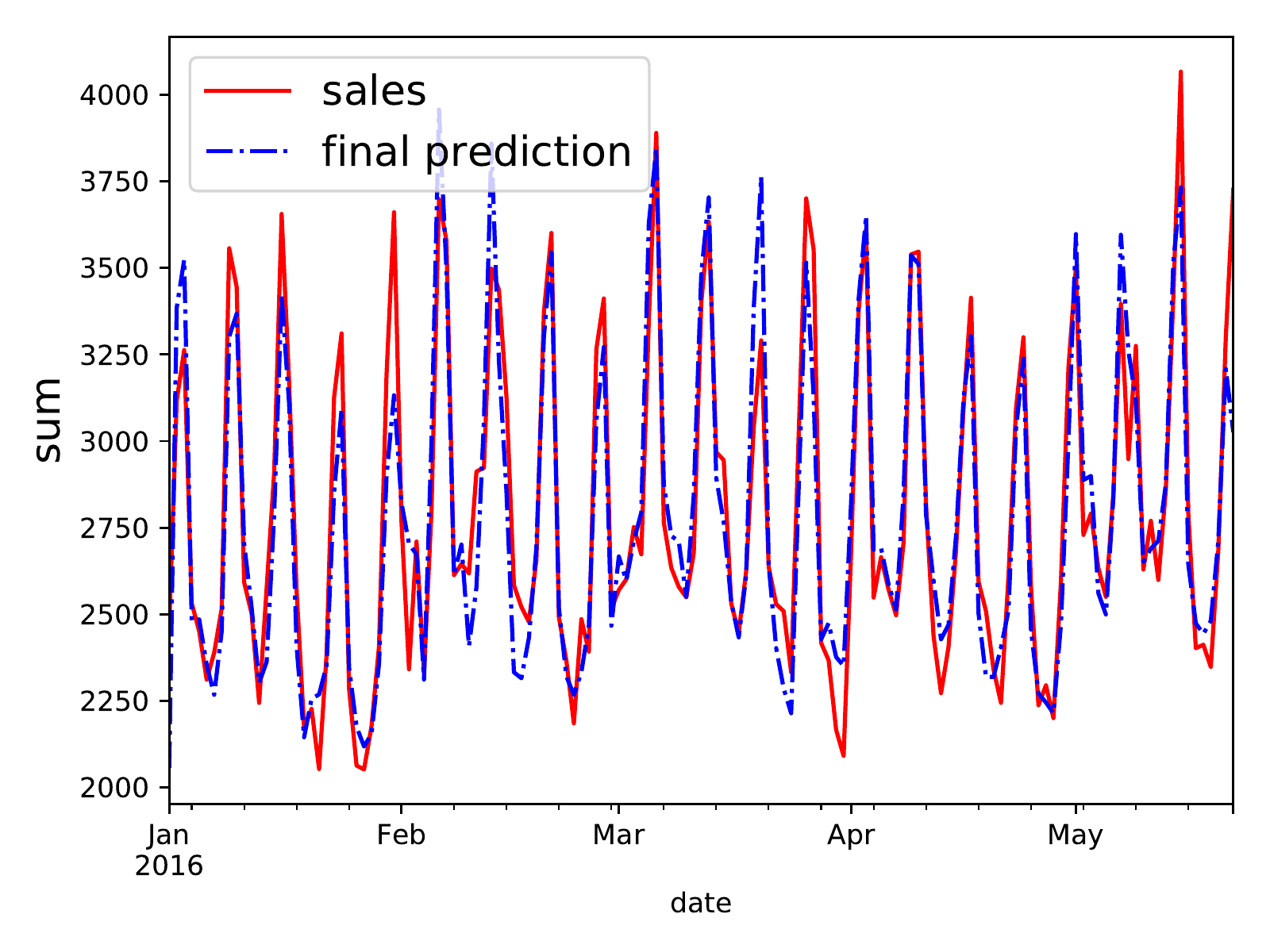}
\caption{\label{fig:mean_prediction} Time series of final mean predictions as well as sales in test period summed over all products and stores.}
\end{center}
\end{figure}

We use the MAD as well as the MSE as two common metrics for the evaluation of point estimates to give a rough estimate of the accuracy of the predicted mean. These metrics do not take the shape of the underlying PDF into account, but only compare the predicted mean to the observed number of sales. As shown in Appendix \ref{sec:CostQuantile}, the mean of a predicted PDF is the optimal point estimator for the MSE and the median for the MAD. However, as we are only interested in relative numbers for the sake of our comparisons here, we just use our mean prediction in both metrics. Using the independent test data, we obtain the metrics as stated in the top line of \tab \ref{tab:mad_mse}. The mean of the target, i.e., the observed sales, is $3.28$ for this period. \fig \ref{fig:mean_prediction} shows the time series of both predictions and sales summed over all 100 products and 10 stores during the test period.

\begin{table}[h!]
\begin{center}
\caption{Accuracy metrics for the predicted mean values for three different model setups: ML model without lagged target information but subsequent residual correction (a), ML model including lagged target information by means of two EMWA features without residual correction (b), and ML model from b) with residual correction (c).}
\label{tab:mad_mse}
\begin{tabular}{c|p{4cm}|c|c}
 & & \textbf{MAD} & \textbf{MSE} \\
\hline
a) & no EWMA features,\newline residual correction & 1.65 & 10.09 \\
\hline
b) & EWMA features,\newline no residual correction & 1.69 & 10.84 \\
\hline
c) & EWMA features,\newline residual correction & 1.68 & 10.41
\end{tabular}
\end{center}
\end{table}

\begin{figure}
\begin{center}
\includegraphics[width=8cm]{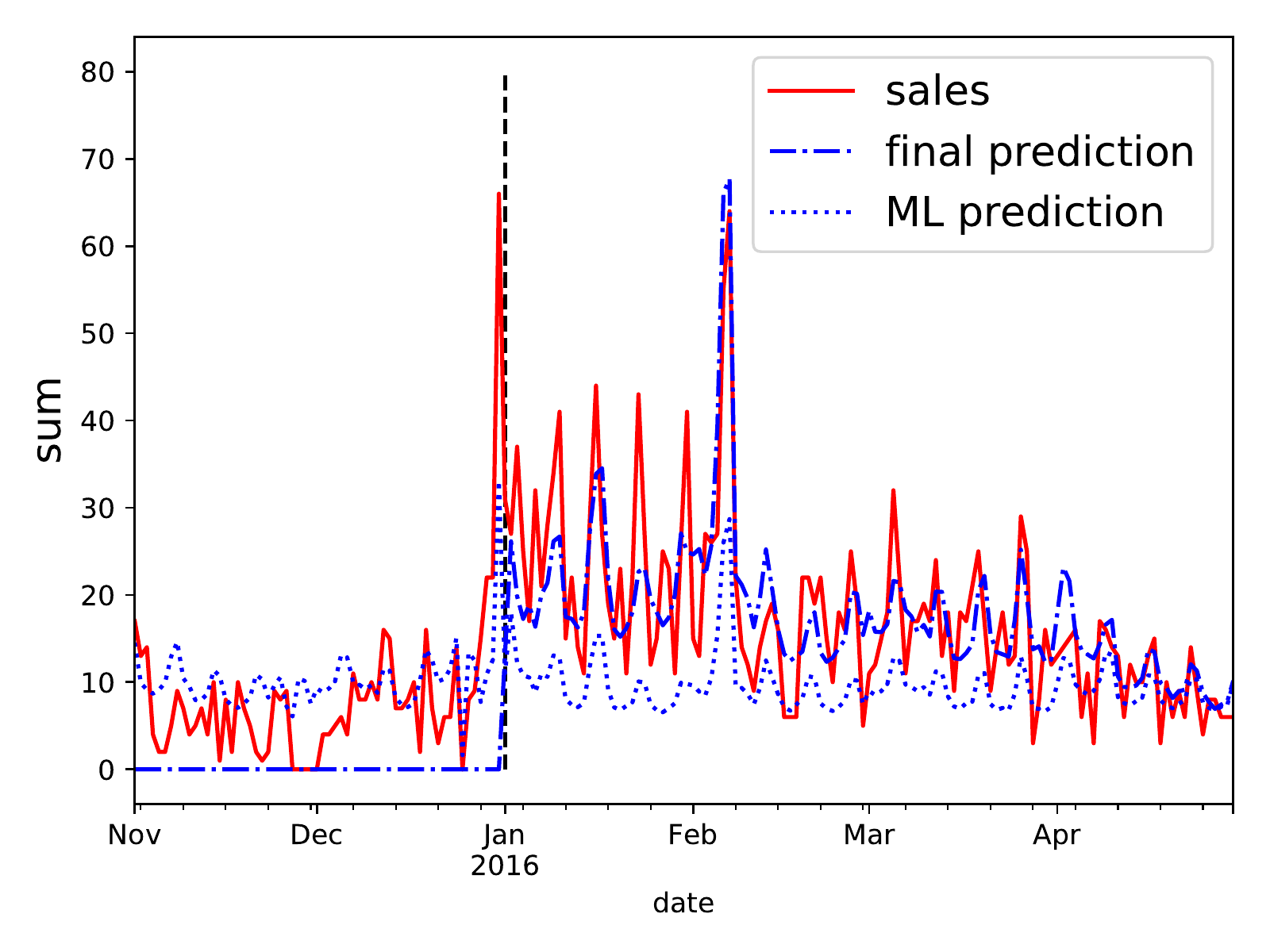}
\caption{\label{fig:ts_res} Time series of mean predictions, before (machine learning prediction) and after (final prediction) residual correction, as well as sales for a specific product-location combination. In addition to the test period in 2016, the last two months of the training period are shown as well (the dashed vertical line marking the transition from training to test period), in order to better demonstrate the workings of the residual correction.}
\end{center}
\end{figure}

To visualize the operation principle of the residual correction method, \fig \ref{fig:ts_res} shows the time series of sales and machine learning (before residual correction) as well as final predictions (after residual correction) for item \texttt{FOODS\_3\_516} in store \texttt{CA\_3} from beginning of November 2015 to end of April 2016, including all days of the test period in 2016 as well as the last two months of the training period before. Hereby, the machine learning predictions in the training period are in-sample predictions, which are just used to calculate the residual correction. A sudden change of the sales time series at the turn of the year is visible, which is seemingly not anticipated by the machine learning model, as its predictions are clearly off in the test period in 2016. However, the residual correction is able to quickly account for the resulting deviations between sales and machine learning predictions, which can be seen by the much better agreement of the final predictions.

\noindent
To give an indication of the effectiveness of the two-step approach, i.e., a machine learning model without lagged target information and residual correction as outlined in sec. \ref{sec:ts_confounding}, we compare its results with a model following the common practice of including the lagged target information as a variety of EMWA variables used as features in \tab \ref{tab:mad_mse} (for the same training and test setup). For this, we include two EWMA features in the machine learning model, one grouped by \texttt{store\_id} and \texttt{item\_id} (with a lag of two days and $\alpha=0.25$), and one grouped by \texttt{store\_id}, \texttt{item\_id}, and the day of the week (with a lag of one week and $\alpha=0.05$), where the values for $\alpha$ are again found empirically. In \tab \ref{tab:mad_mse}, we also show the results after applying the same residual correction as before for the machine learning model without lagged target information on top of this model including the two EWMA features. The crucial point in this is that when applying the same residual correction method to both machine learning models with and without including lagged target information as EWMA features, the one without yields better results. Following our reasoning in sec. \ref{sec:ts_confounding}, this is due to detrimental temporal confounding effects disturbing the learning of the causal dependencies between the exogenous features and the target, which in turn leads to a worse generalizability of the machine learning model.

\subsection{Variance Estimation}

The second model, based on Cyclic Boosting in its negative binomial width mode described in sec. \ref{sec:cb_width}, is used to estimate the variance of the NBD. The data are split into training and test sets as above, and in addition, the mean predictions for each individual product-location-date combination are fixed in the variance model as stated by \eqn \eqref{eqn:loss_likelihood}. This effectively means that the needed mean predictions are created in-sample for the training period using the fully trained and validated model for the mean discussed above.

\begin{figure}
\begin{center}
\includegraphics[width=4cm]{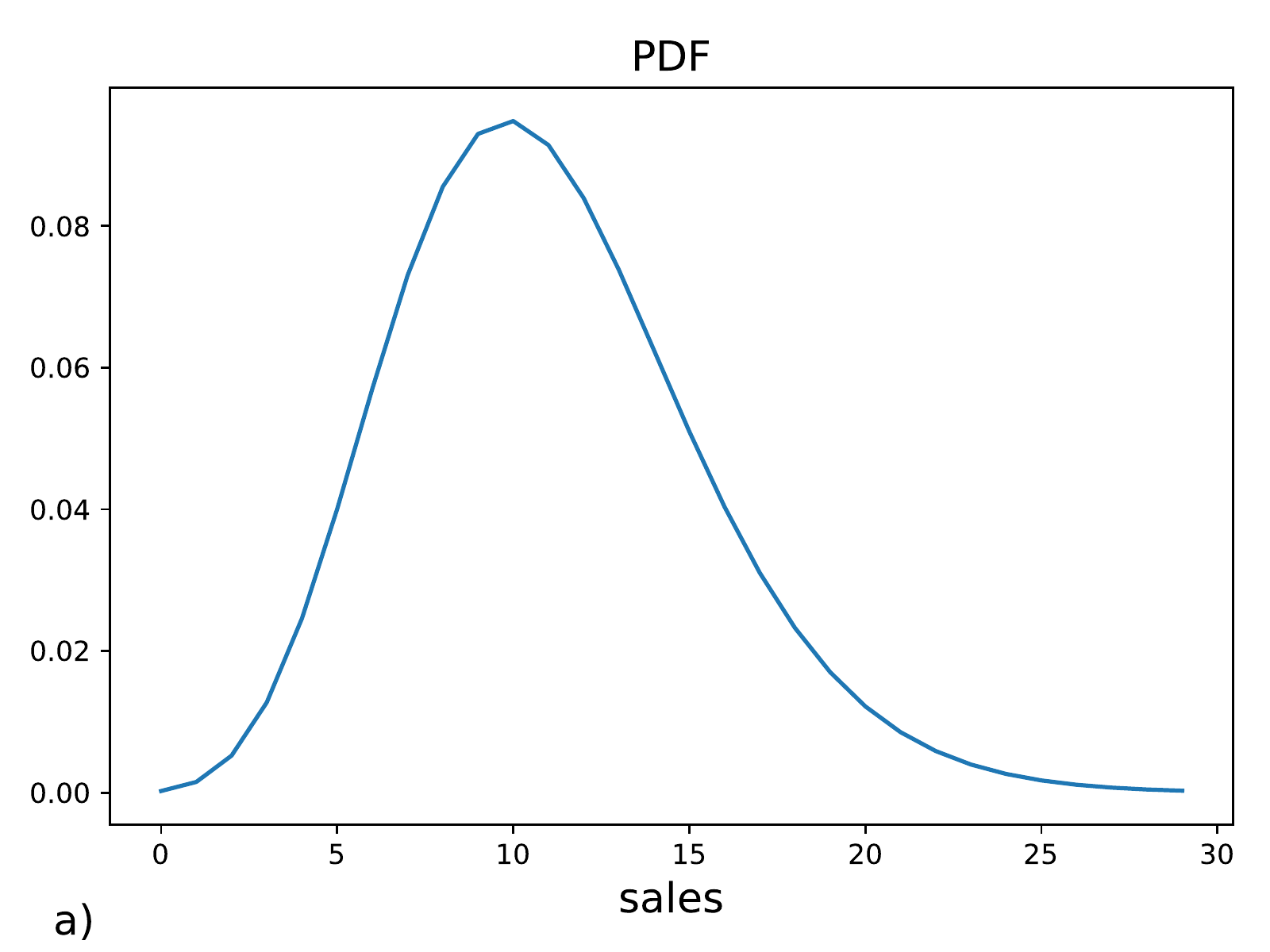}
\includegraphics[width=4cm]{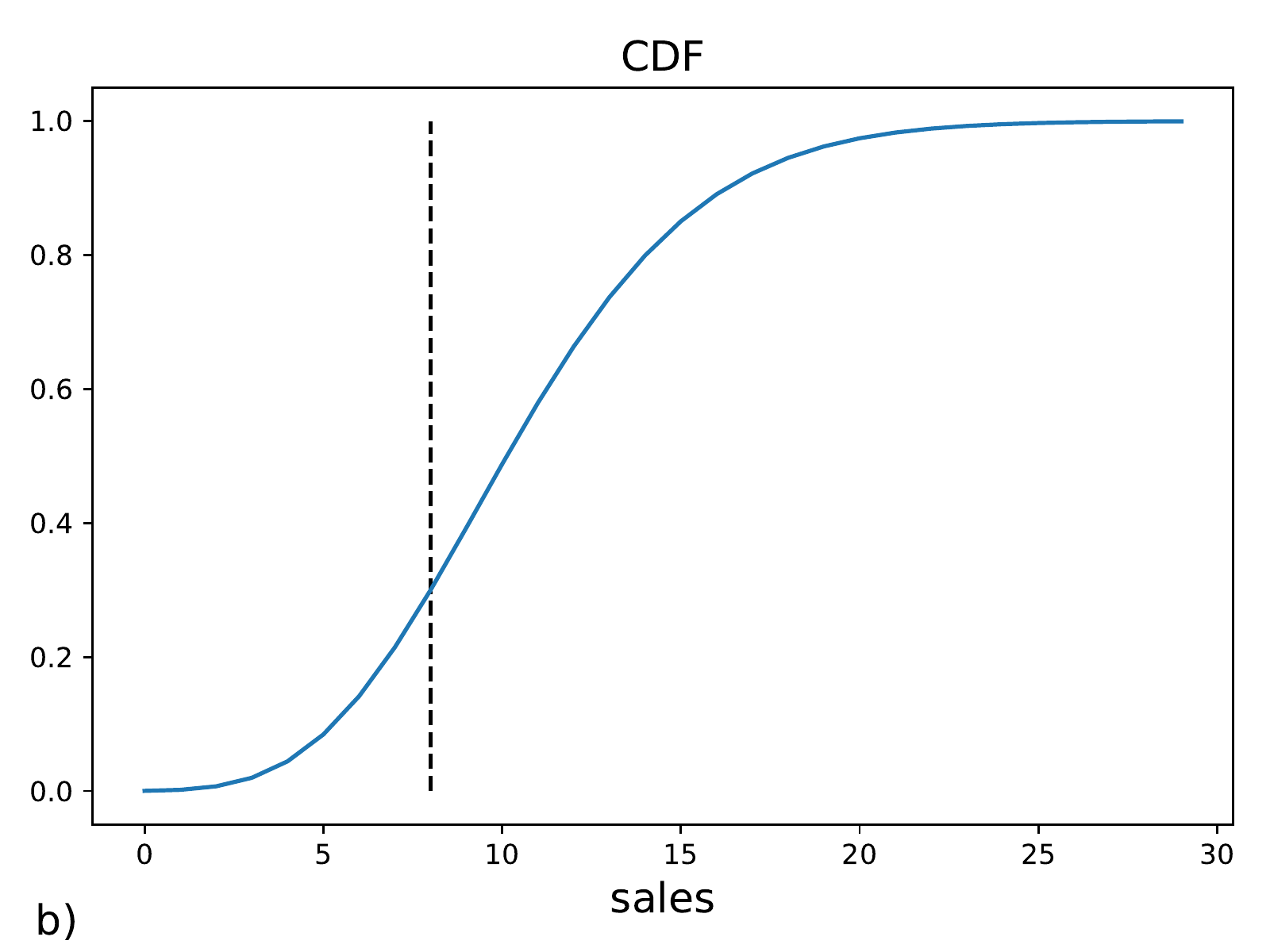}
\caption{\label{fig:pdf_example} Predicted PDF (a) and CDF (b) for a specific product-location-date combination. The dashed vertical line in the CDF plot represents the actually observed sales, corresponding to a CDF value just under $0.3$.}
\end{center}
\end{figure}

\noindent
In this model focusing on the variance, we use the same set of features as for the mean model described above in sec. \ref{sec:example_mean}, except for dropping most of the two-dimensional combinations (only keeping \texttt{item\_id} - \texttt{store\_id}, \texttt{store\_id} - day of week, and \texttt{item\_id} - event type features) and adding the in-sample mean prediction as feature. An example for the resulting PDF and CDF predictions for item \texttt{FOODS\_3\_516} in store \texttt{TX\_3} on 2016-05-06, together with the corresponding observed sales value, is shown in \fig \ref{fig:pdf_example}.

\begin{figure}
\begin{center}
\includegraphics[width=4cm]{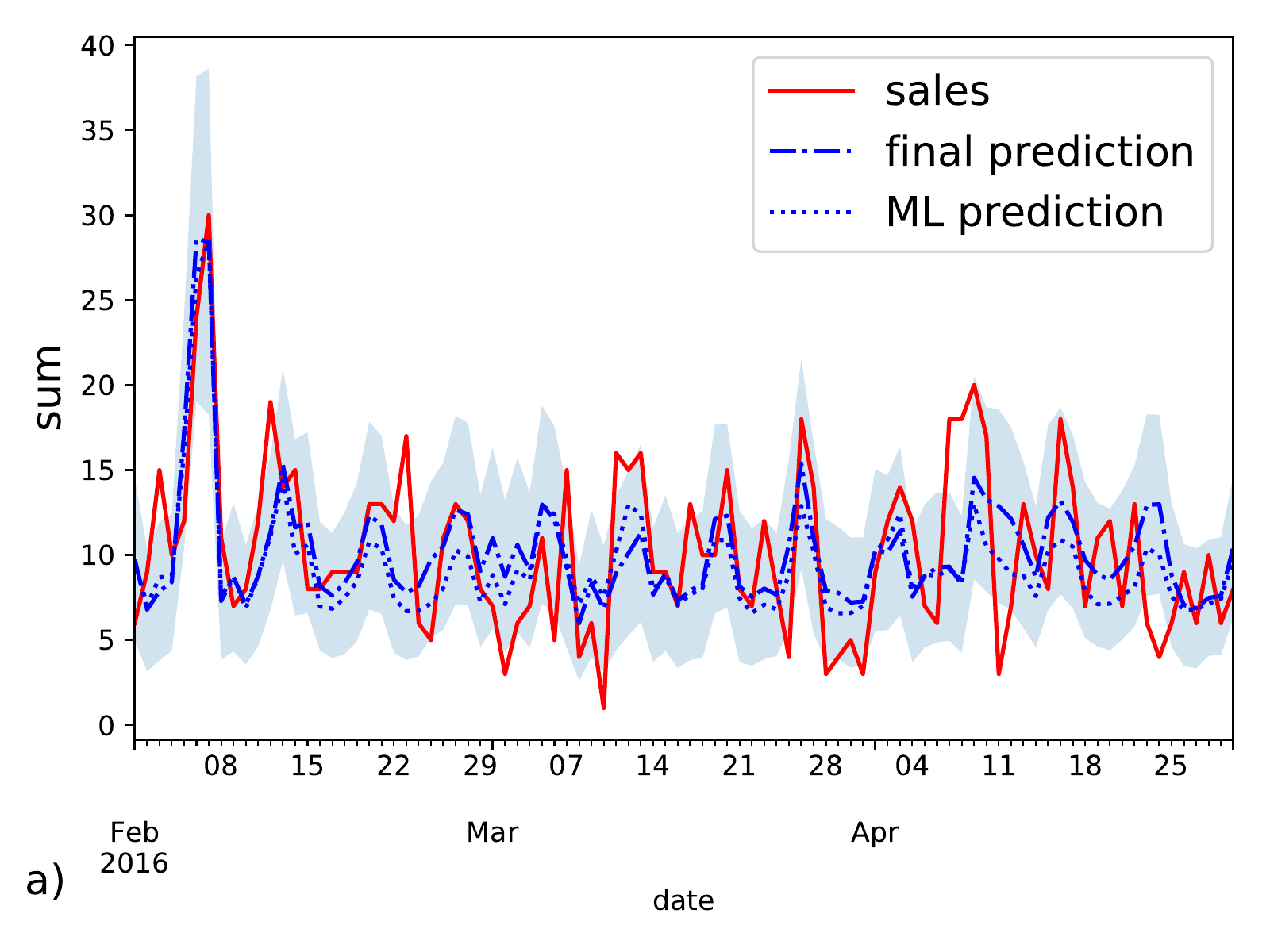}
\includegraphics[width=4cm]{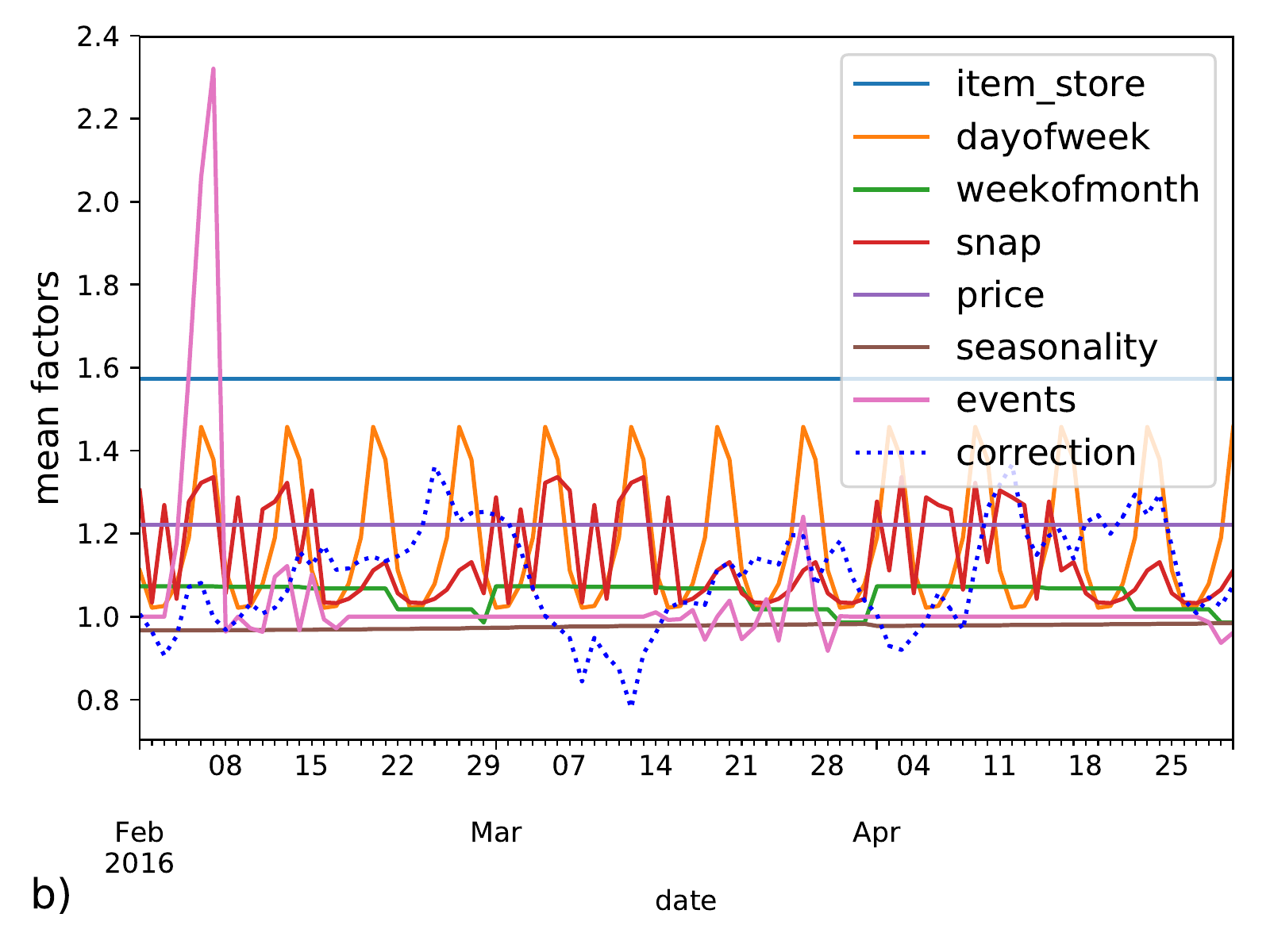}
\includegraphics[width=4cm]{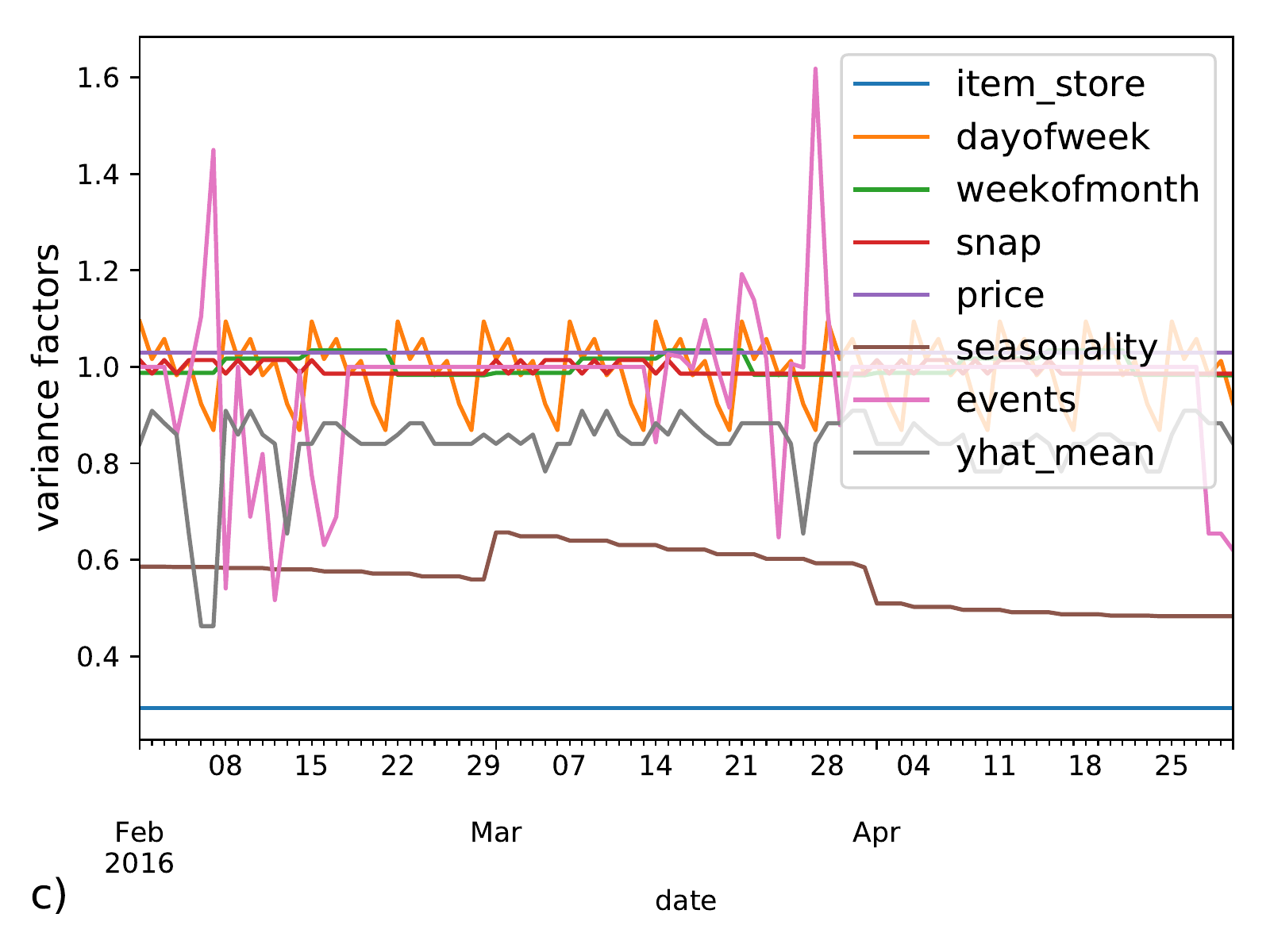}
\caption{\label{fig:factor_plots} Time series of mean predictions (with the surrounding transparent band simplistically indicating a standard deviation as square root of the corresponding variance predictions), before (machine learning prediction) and after residual correction (final prediction), as well as sales (a), constituting factors of the mean predictions (including the factor of the residual correction) (b), and constituting factors of the variance predictions (c), for a specific product-location combination from beginning of February to end of April 2016. For both mean and variance predictions, the most granular constituting factors are aggregated, i.e., multiplied, for the sake of visualization.}
\end{center}
\end{figure}

\fig \ref{fig:factor_plots} visualizes the fully explainable nature of the individual Cyclic Boosting predictions, both for the mean and the variance model, by showing the time series plots of the factor values constituting the corresponding predictions for item \texttt{FOODS\_3\_516} in store \texttt{TX\_3} for all days in the test period from beginning of February to end of April 2016. Hereby, factors of different individual features, as described above, are multiplied, according to \eqn \eqref{eqn:cb} for the mean predictions and \eqn \eqref{eqn:r} for the variance predictions, in order to represent the behavior of a feature group. For example, the \texttt{events} feature group corresponds to the combination of all the factors of the different event features and the \texttt{dayofweek} feature group includes the one-dimensional feature \texttt{dayofweek} and its two-dimensional combinations with \texttt{store\_id} and \texttt{item\_id}. Furthermore, the values of the residual correction on the mean predictions are included, which correspond to the differences between the shown machine learning and final predictions.

\noindent
The different \texttt{item\_store} values for mean (higher than $1$) and variance model factors (lower than $1$), including the static one-dimensional features \texttt{store\_id} and \texttt{item\_id} and their two-dimensional combination, represent the fact that this product-location combination sells more than the average of all product-location combinations, and the inverse dispersion parameter $1/r$ in \eqn \eqref{eqn:variance_r} in turn tends toward lower values. The explanation of the model for the peak in sales and mean prediction on February 7, 2016 is the event Super Bowl. The variance prediction for this day is driven by two competing factors: the Super Bowl event itself drives the variance up, while the high mean prediction (as feature) drives it down.

\subsection{Evaluation of PDF Predictions}

\begin{figure}
\begin{center}
\includegraphics[width=4cm]{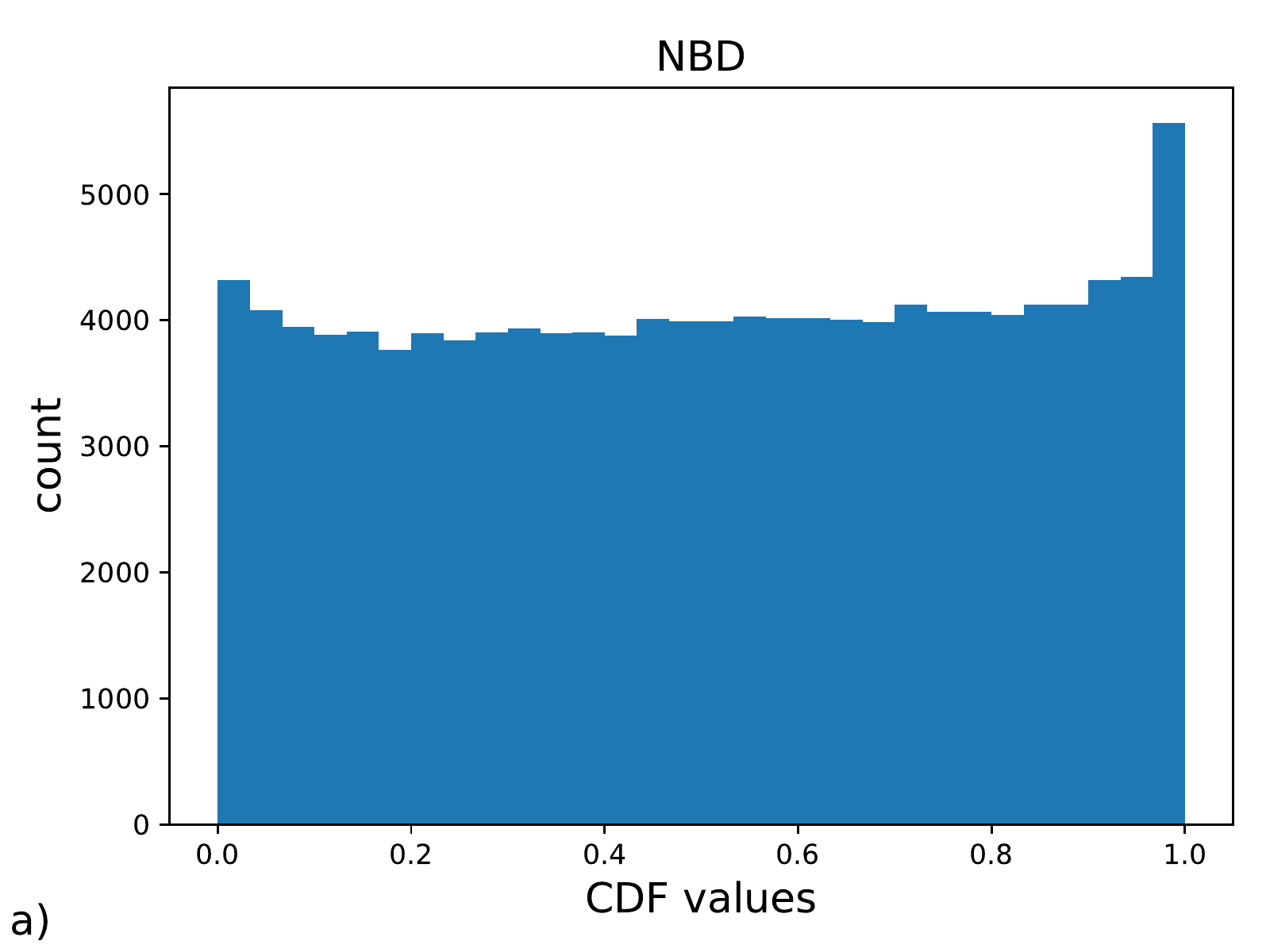}
\includegraphics[width=4cm]{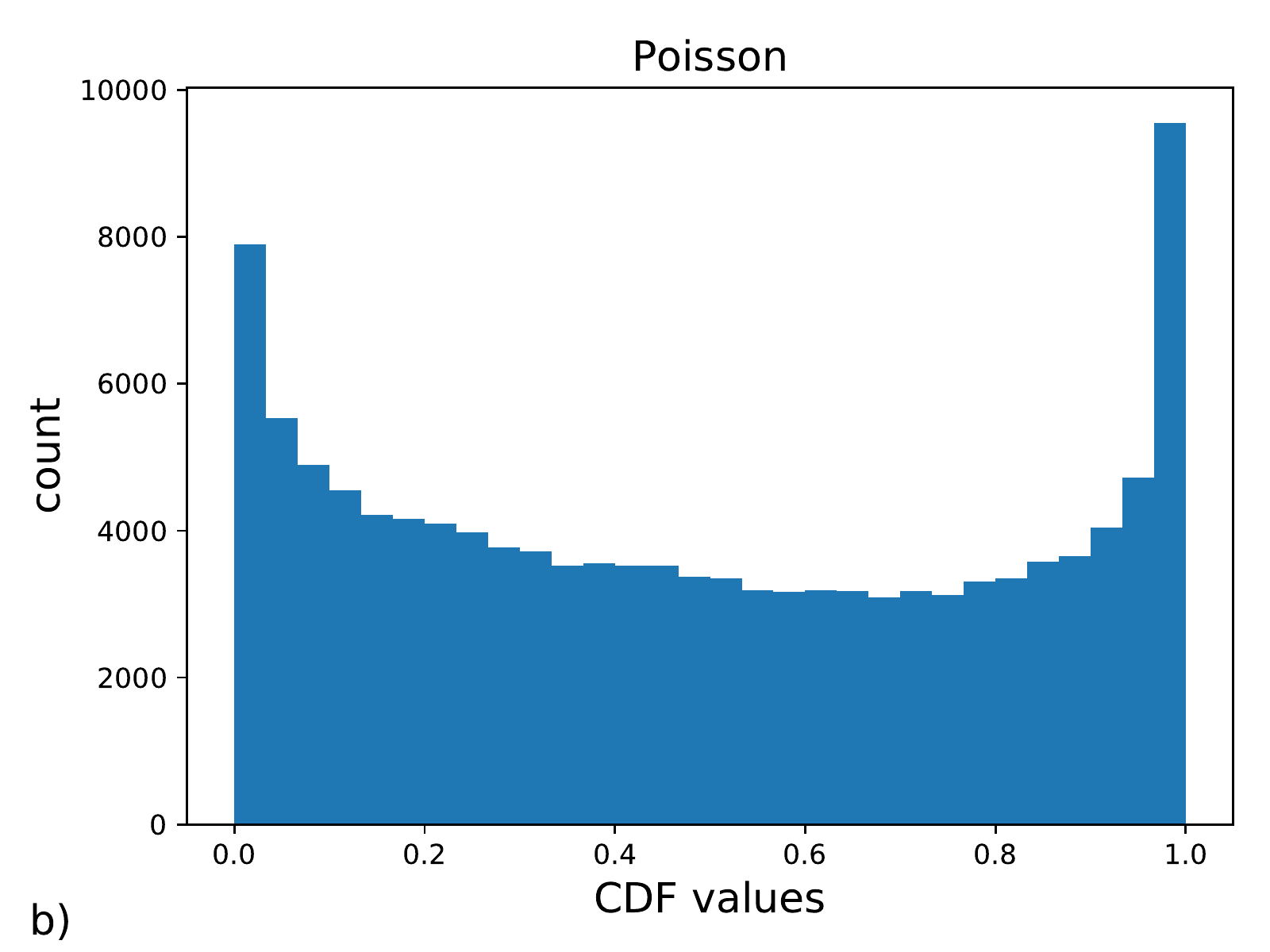}
\includegraphics[width=4cm]{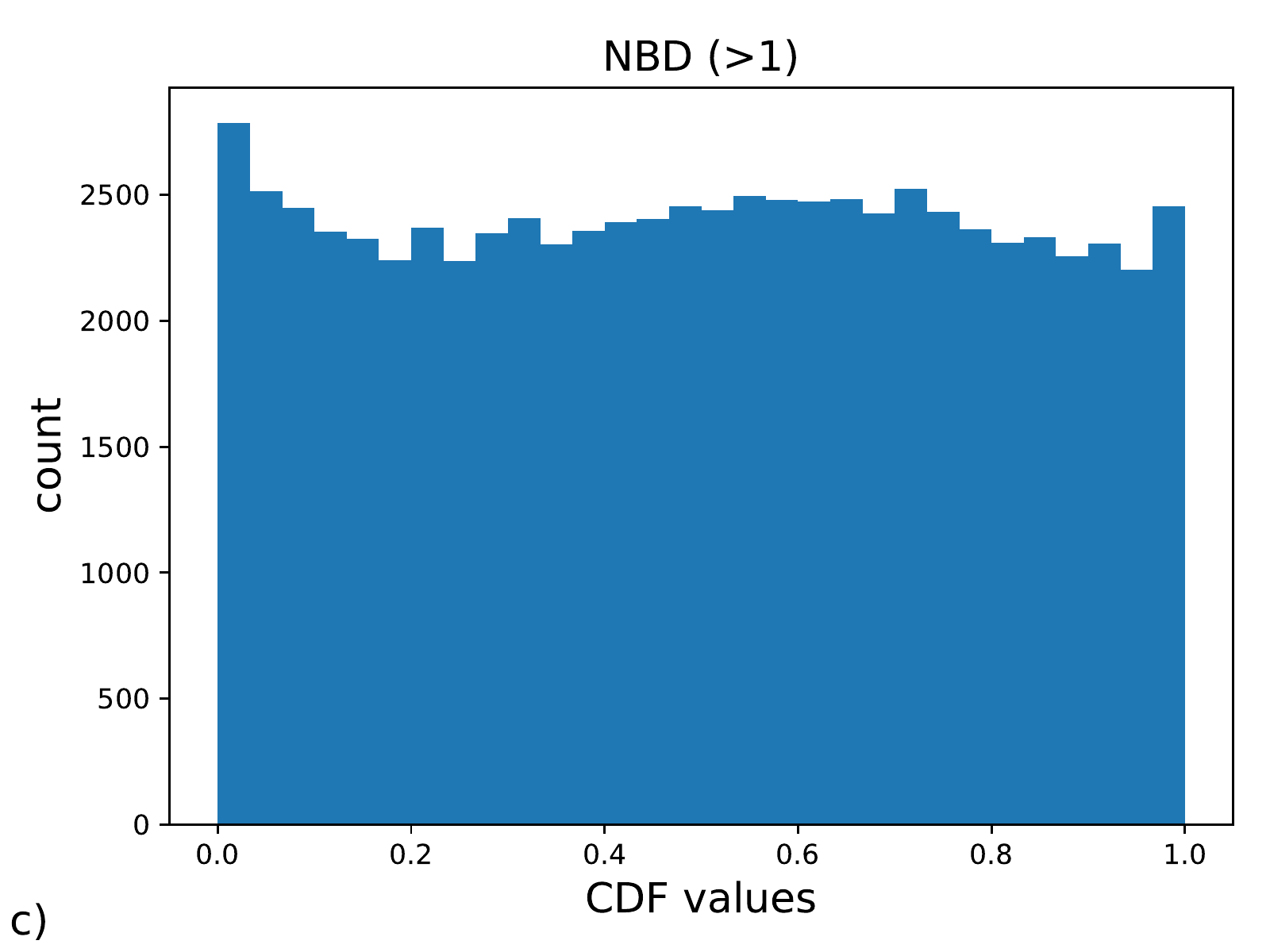}
\includegraphics[width=4cm]{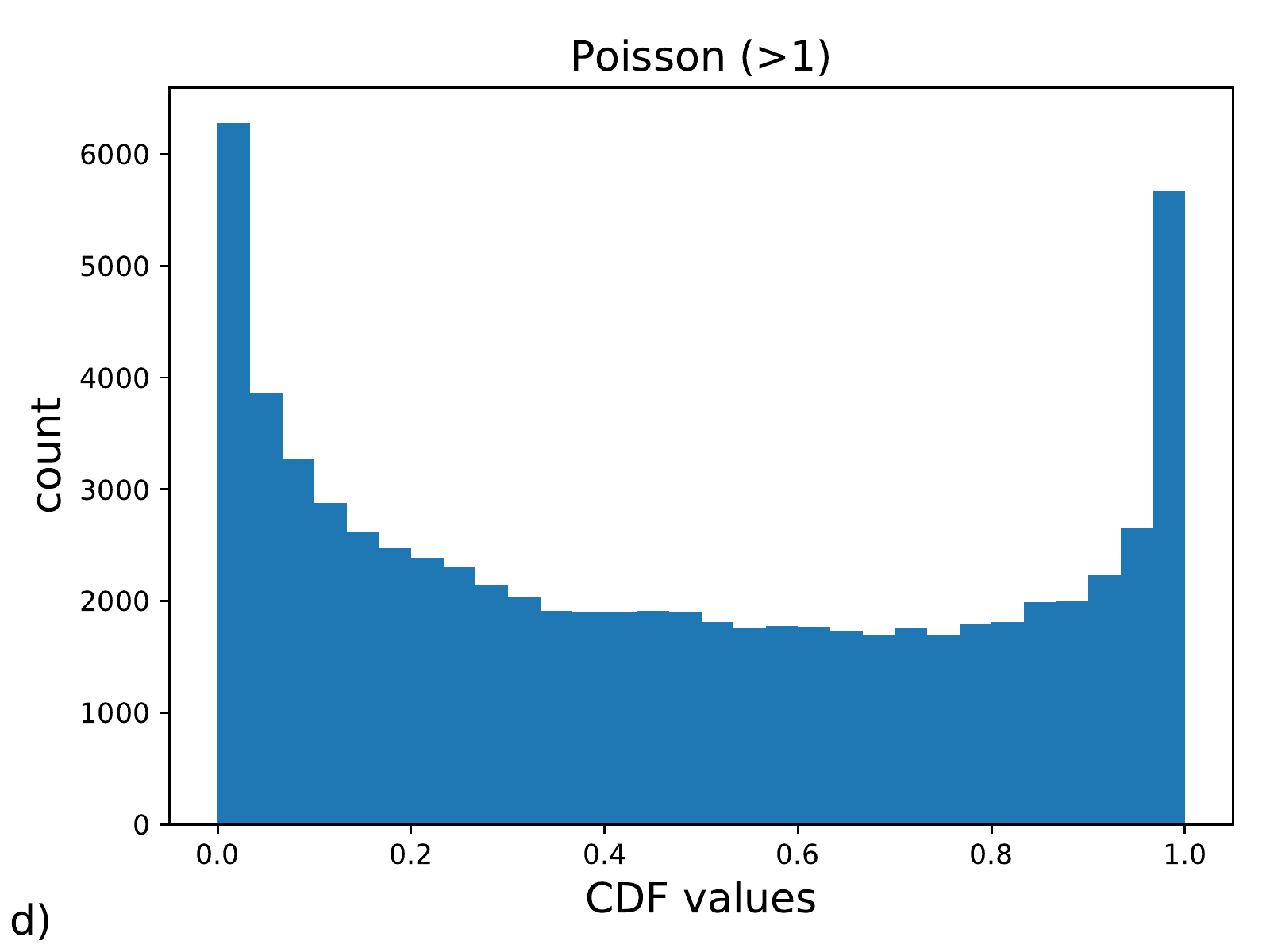}
\caption{\label{fig:cdf_demand} Histograms of ex post observed target CDF values of the corresponding individual PDF predictions (to be compared to a uniform distribution) for our negative binomial model (a) and a simpler Poisson model for comparison (b), using all product-location-day combinations in the test period. In order to show the effect of slow-sellers, both negative binomial (c) and Poisson model (d) histograms are also shown using samples with mean predictions higher than $1.0$ only.}
\end{center}
\end{figure}

\fig \ref{fig:cdf_demand} shows the histogram of CDF observations according to the method described in sec. \ref{sec:cdf_histo} for all product-location-day combinations in the test period. As benchmark, we compare the outcome of our negative binomial model to a simpler Poisson assumption, which has only a single model parameter, the mean. Using the same mean predictions for both negative binomial and Poisson model, the negative binomial PDF predictions are much closer to the uniform distribution, which we expect for optimal PDF predictions, than the Poisson PDF predictions, showing the effectiveness of our variance estimation. While the Poisson histogram shows a clear pattern of too narrow PDF predictions (see \fig \ref{fig:cdf_histos}), the most significant deviations of the negative binomial histogram from the uniform distribution can be found in the first bins of CDF values close to $0$ (overprediction) as well as in the last bins of CDF values close to $1$ (underprediction). The first case is mainly due to a slight zero-inflation of actual sales. The latter case primarily stems from slow moving articles with less than one item sold per store per day on average, what can be seen in plots c and d of \fig \ref{fig:cdf_demand}, where the spike at CDF values close to $1$ is reduced by excluding all samples with mean predictions lower than $1.0$. The resulting distribution for the negative binomial model indicates a slight tendency toward too broad PDF predictions, at least for some observations.

\begin{figure}
\begin{center}
\includegraphics[width=4cm]{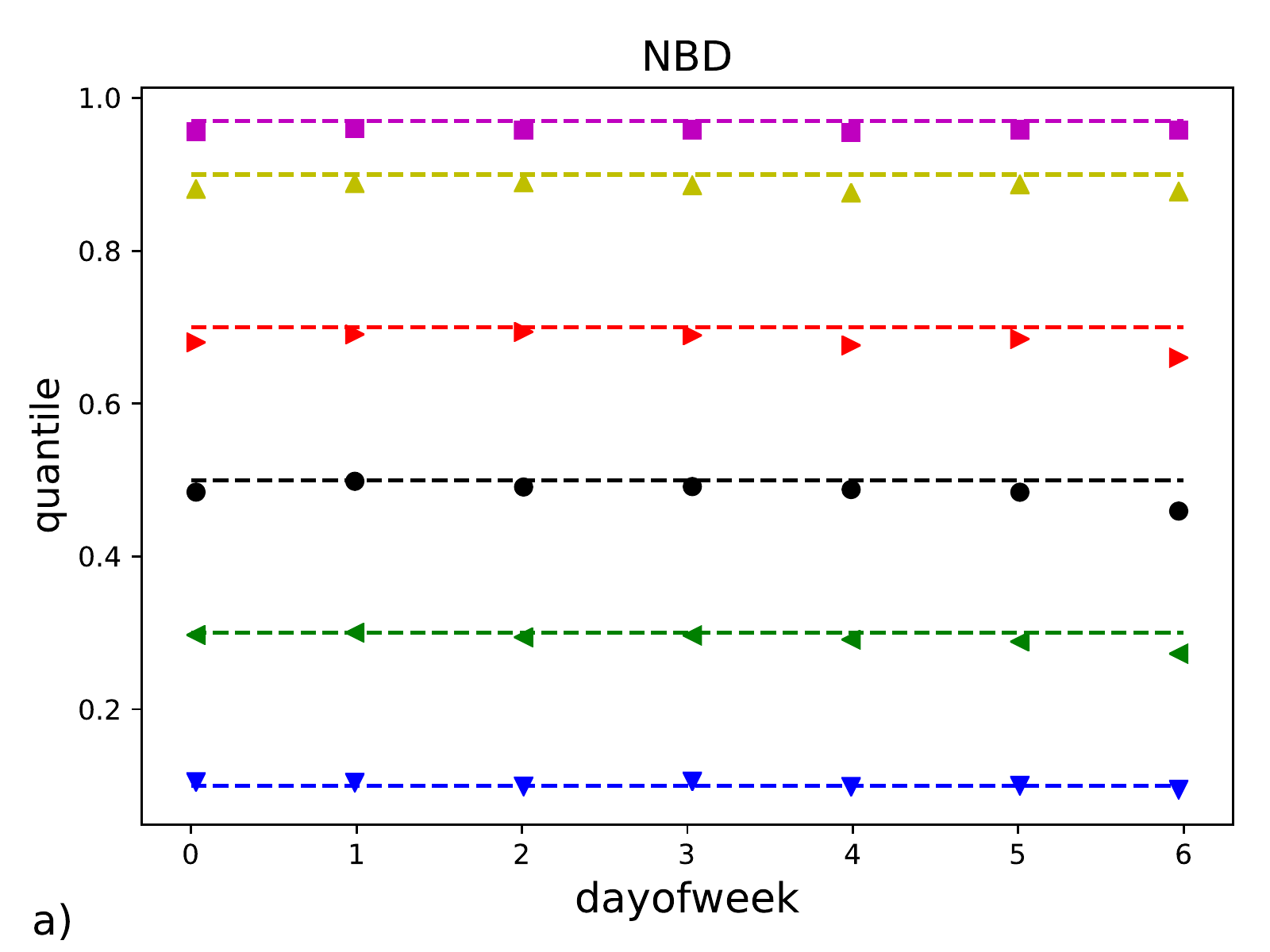}
\includegraphics[width=4cm]{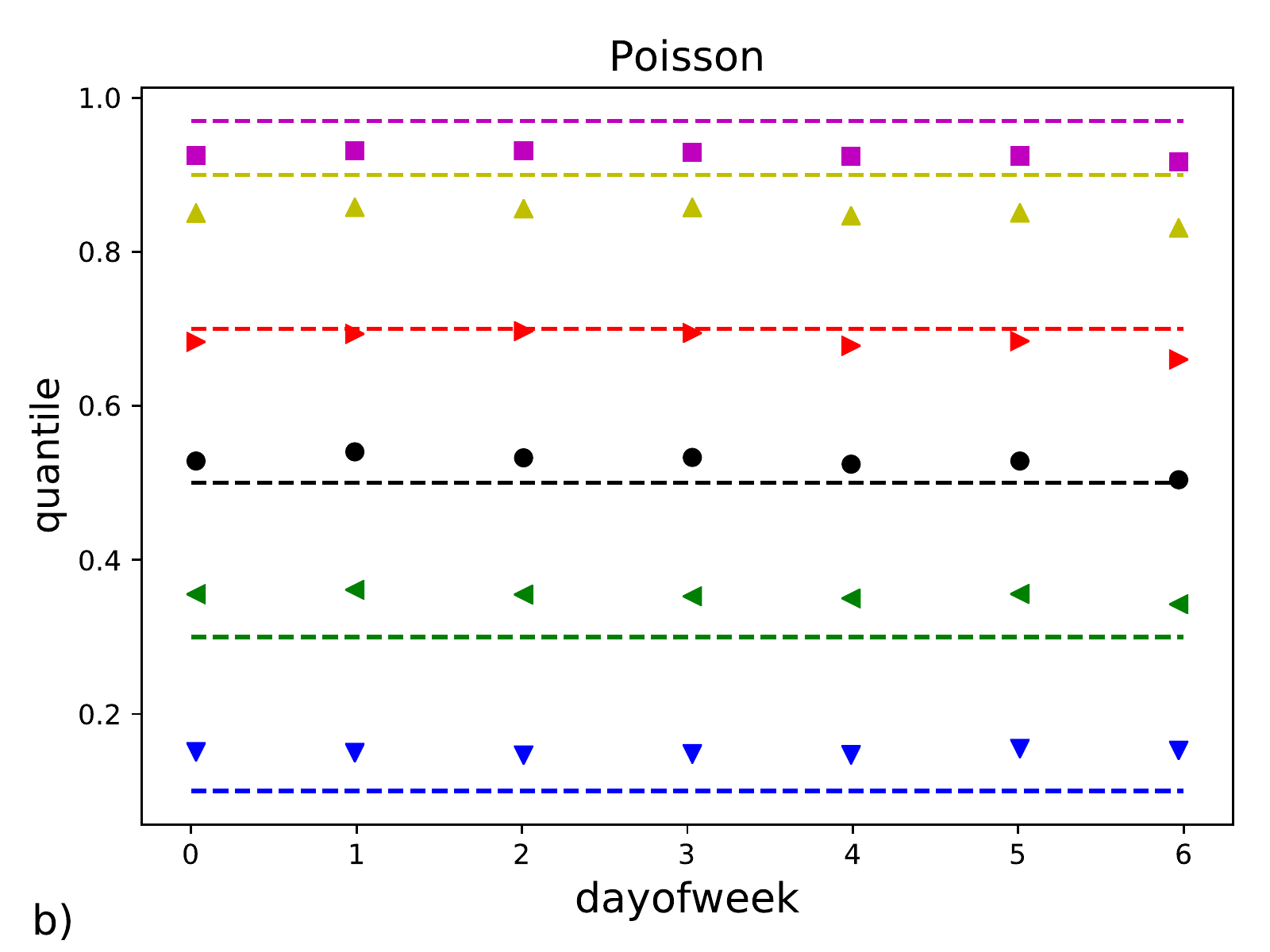}
\caption{\label{fig:invquant_dayofweek} Inverse quantile profile plot for the different days of the week on the $x$ axis (from Monday to Sunday) aggregated over all product-location-day combinations in the test period, for our negative binomial model (a) and a simpler Poisson model for comparison (b).}
\end{center}
\end{figure}

Using the inverse quantile profile plots introduced in sec. \ref{sec:invquant_plot}, the prediction quality of the full predicted PDF can be assessed in more detail. In \fig \ref{fig:invquant_dayofweek}, each column corresponds to a day of the week, using 0 for Mondays, 1 for Tuesdays, etc., and the considered quantiles are 0.1, 0.3, 0.5, 0.7, 0.9, and 0.97. Again, we compare the outcome of our negative binomial model to a simpler Poisson assumption. Aggregating over all stores, items, and sales dates (independently for each day of the week), we can see that for the negative binomial model each quantile is predicted relatively well, except for Sundays showing a tendency for overprediction of the mean (see \fig \ref{fig:invquant_example}). The Poisson model, on the other hand, shows a significant pattern of too narrow PDF predictions across all days of the week.

\begin{figure}
\begin{center}
\includegraphics[width=4cm]{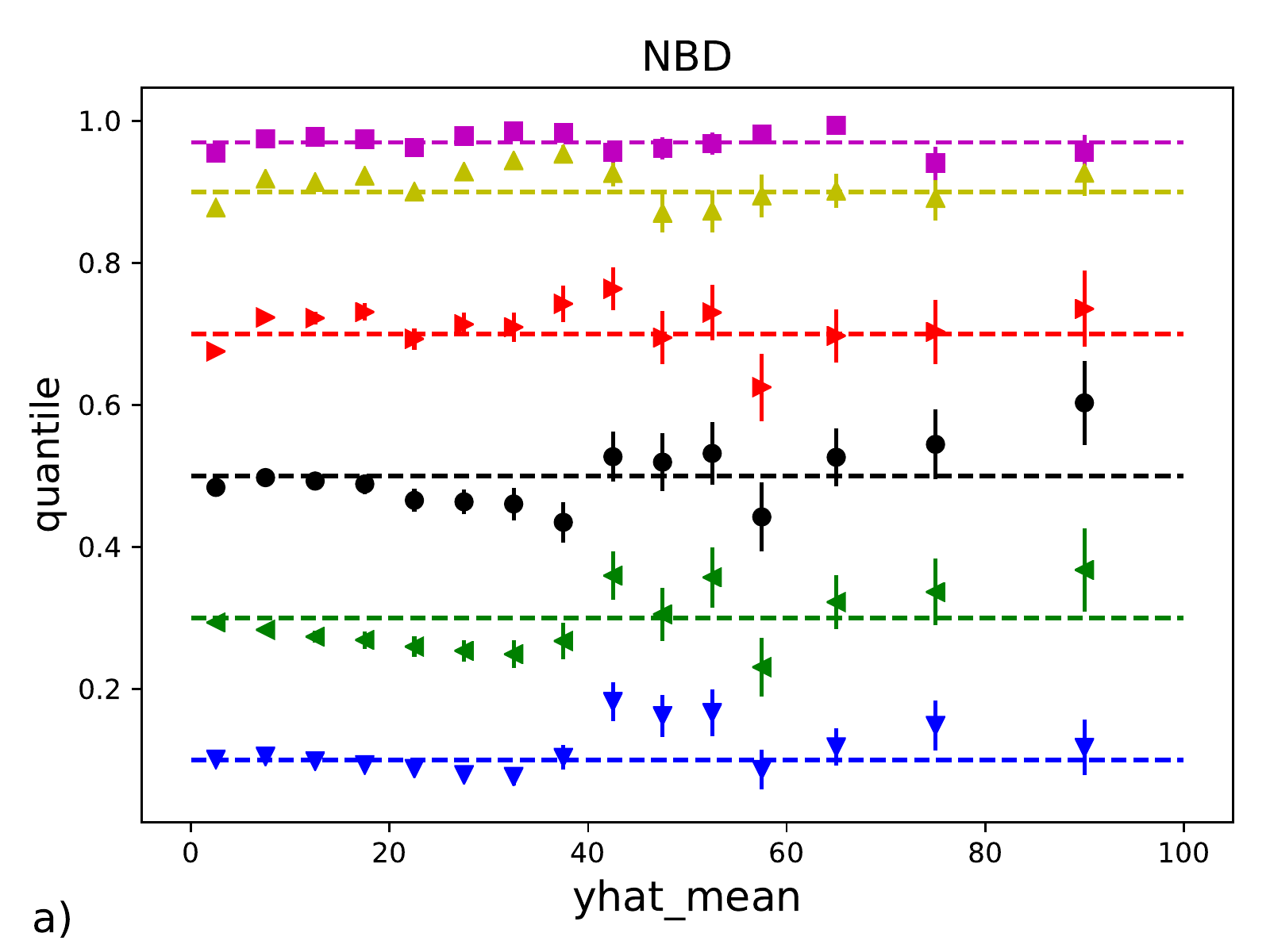}
\includegraphics[width=4cm]{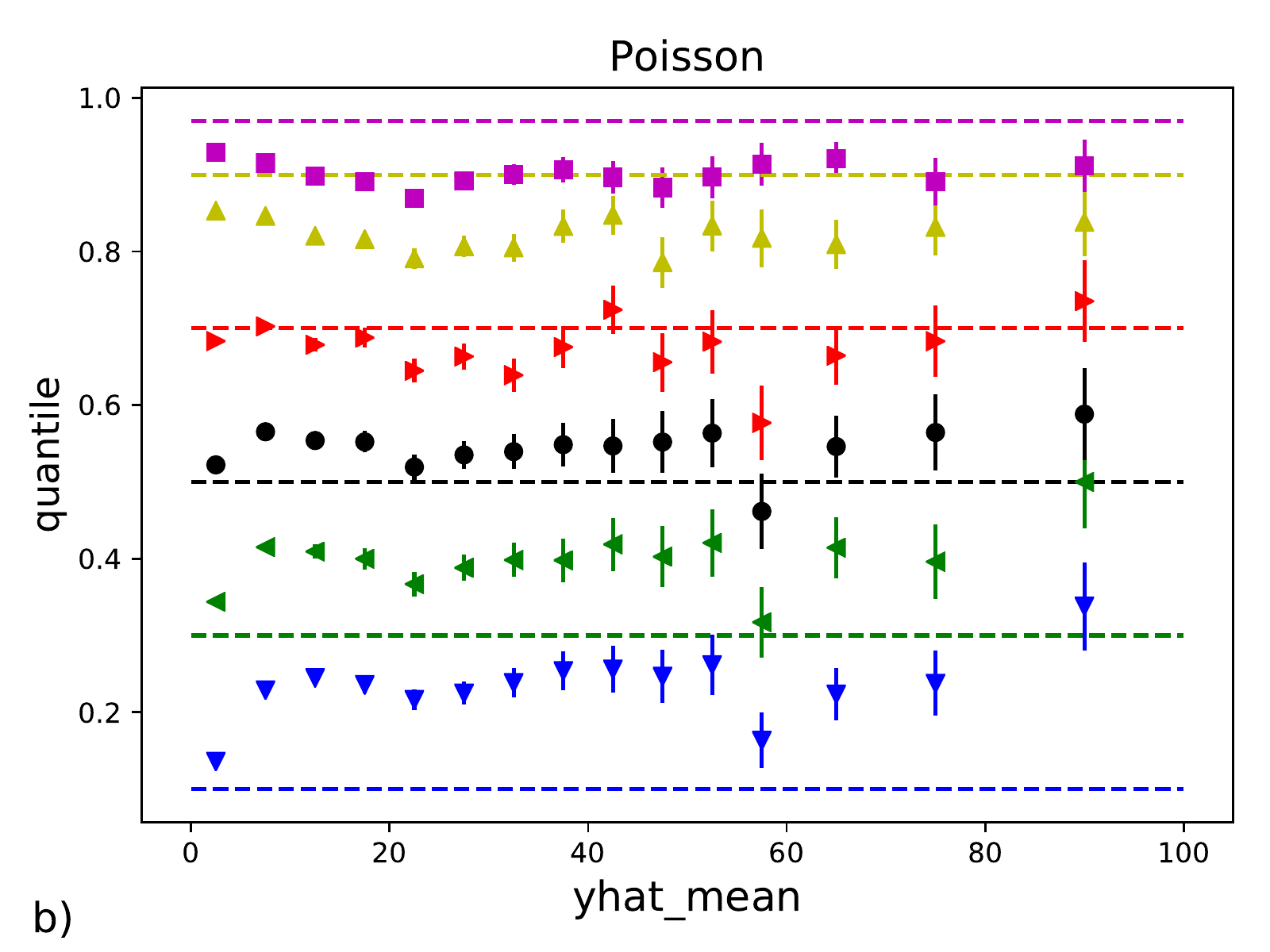}
\caption{\label{fig:invquant_mean} Inverse quantile profile plot for mean predictions on the $x$ axis aggregated over all product-location-day combinations in the test period, for our negative binomial model (a) and a simpler Poisson model for comparison (b).}
\end{center}
\end{figure}

\fig \ref{fig:invquant_mean} shows the inverse quantile profile plot using the predicted mean as the $x$ axis of the graph, meaning that the distributions are grouped such that the 15 columns correspond to mean predictions in 12 intervals with widths of 5 from 0.0 to 60.0 and 3 remaining intervals $[60.0, 70.0]$, $(70.0, 80.0]$, and $(80.0, 100.0]$ (with mean predictions higher than 100.0 included in the highest interval), while we aggregate over all locations, items, and sales dates. The considered quantiles are chosen to be the same as in \fig \ref{fig:invquant_dayofweek}. While the Poisson model again shows a significant pattern of too narrow PDF predictions across the full range of mean prediction values, we can also see several deviations from the expected uniform behavior for the negative binomial model. For mean predictions around $30$, the negative binomial model deviates significantly from the expected behavior, with the shape of the deviations pointing toward too broad PDF predictions (as also observed in the CDF histogram above). And for mean predictions higher than $80$, the pattern of the deviations indicate an underprediction of the mean parameter. The deviations at very low predicted mean values reflect the complications of zero-inflation and slow-sellers seen in the first and last bins of the CDF histograms above.

\noindent
In a real-live situation with a live supply chain project for a customer, a more detailed investigation of the root-causes would then start. However, it should be noted that due to this segregation, the statistics in each part of the relevant test sample becomes a limiting factor as well. This plot also illustrates the benefits of using the quantile profile plot, since we have seen that using more conventional approaches, such as \fig \ref{fig:mean_prediction} or even \fig \ref{fig:cdf_demand}, the predictions appear reasonable, even when not relying on simple point metrics such as MAD or MSE. It is therefore paramount both to predict the full PDF instead of just a point estimate, such as the mean, as well as verifying a number of quantiles for each prediction to assess the quality of the prediction thoroughly.

The quantitative results for the CDF accuracy of our PDF predictions using the first Wasserstein distance as metric (as described in sec. \ref{sec:cdf_acc}) and calculated over all product-location-day combinations in the test period, can be found in \tab \ref{tab:cdf_acc}. As benchmark, we again compare against a Poisson model assumption using the same mean predictions. As expected from the qualitative findings in \fig \ref{fig:cdf_demand}, the negative binomial PDF predictions show a significant improvement over the simpler Poisson model.

\begin{table}[h!]
\begin{center}
\caption{Accuracy for negative binomial and Poisson PDF predictions, using the first Wasserstein distance as metric, calculated over all product-location-day combinations in the test period.}
\label{tab:cdf_acc}
\begin{tabular}{c|c|c}
 & \textbf{NBD} & \textbf{Poisson} \\
\hline
\textbf{EMD accuracy} & 0.967 & 0.850
\end{tabular}
\end{center}
\end{table}

\section{Conclusion}
Demand forecasting remains a crucial step in operational planning for retailers. Both from practical and theoretical perspectives, disentangling the forecast of demand from the operational decision making regarding order quantities has potentially significant advantages over an integrated approach using data to predict the resulting order quantities directly. When predicting future demand in the form of individual probability distributions, both model-free and model-based approaches can be used, where the model-based approach is generally more robust, as it is rooted in a theoretical understanding of the sales process. Compared to a model-free approach like quantile regression, the distributional assumption can drastically reduce the uncertainty of the resulting predictions. Using the Cyclic Boosting supervised machine learning approach, the full probability density function can be predicted in a fully explainable way on the individual level. This allows to extract not only simple point estimates, such as the expected mean demand, but all relevant quantiles of the distribution.

Once the full individual probability distributions are forecasted, their evaluation poses significant challenges: Common metrics, such as the mean absolute deviation or mean squared error, only take point estimates into account and are generally not suitable for the evaluation of predicted PDFs. Furthermore, since the distributions are predicted for individual events, statistical techniques aiming at the comparison of distributions using many events cannot be used in general. Instead, novel techniques exploiting the probability integral transform, namely the histogram of observed CDF values of the predicted individual PDFs as well as inverse quantile profile plots, allow a detailed investigation into the behavior of the predictions. Finally, a quantitative assessment resulting in a single number can be obtained using metrics such as the earth mover distance in a comparison between the histogram of observed CDF values and the expected uniform distribution.

An additional challenge for demand forecasting in specific and time series forecasting in general is the handling of the detrimental effects of temporal confounding, especially on explainability and generalizability of the employed models. Using a two-step model consisting of a machine learning model, including all available exogenous variables as features but excluding any lagged target information, to control for all known temporal confounders, and a subsequent empiric residual correction to capture any information on recent trends missed in the machine learning model, temporal confounding can be avoided, or at least mitigated.

% Authors must disclose all relationships or interests that could have direct or potential influence or impart bias on the work: 
%
%\section*{Declarations}
%
%\subsection*{Funding}
%No external funding was obtained for this work.
%
%\subsection*{Conflict of interest}
%
%\noindent
%Authors Felix Wick and Michael Feindt applied for a US patent ``A System and Method of Cyclic Boosting for Explainable Supervised Machine Learning''.
%
%\noindent
%Authors Felix Wick, Martin Hahn, and Moritz Wolf applied for a US patent ``Causal Factor Machine Learning with Individual Negative Binomial Probability Density Function Estimation''.
%
%\noindent
%Authors Felix Wick and Trapti Singhal applied for a US patent ``Evaluation of Predictions as Individual Probability Density Functions''.
%
%\noindent
%Authors Felix Wick and Daniel Stemmer applied for a US patent ``Causal Factor Machine Learning with Residual Time Series Correction''.
%
%
%\subsection*{Availability of data and material}
%Data used in this work are public \cite{kaggle_data}
%
%\subsection*{Code availability}
%The details of the code are available from the authors
%
%\subsection*{Authors' contributions}
%F. Wick: code examples and manuscript text, U. Kerzel: manuscript text, M. Hahn, M. Wolf, T. Singhal, M. Feindt, D. Stemmer, J. Ernst: revision

\bibliography{paper}

\bibliographystyle{ieeetr}

%%%%%%%%%%%%%%%%%%%%%%%%%%%%%%%%%%%%%%%%%%%
%% Appendices
%%%%%%%%%%%%%%%%%%%%%%%%%%%%%%%%%%%%%%%%%%%%
\appendix

\section{Optimal Point Estimator for Given Cost Function}
\label{sec:CostQuantile}

Operational decisions ultimately need to be based on a single number. For example, if we want to order new goods, we need a single number representing the expected demand, and then include operational constraints such as lot sizes and delivery schedules from the manufacturer or wholesaler to arrive at the optimal order quantity for the next delivery period. If we predict the future demand as a complete PDF, we need to determine the optimal point estimator that ``reduces'' the PDF to the single best quantity. In an ideal case, the full cost function $C(p_i,t_i)$ relating point estimator $p_i$ to observed (true) value $t_i$ is available. If the cost function is fully specified, we can obtain the optimal point estimator $p$ by minimizing the cost function with respect to the predicted probability distribution $f(t)$, i.e., we require that the first derivative of the cost function vanishes:
\begin{equation}
	\frac{\partial E[C(p,t)]}{\partial p} = 0
\end{equation}
Since we are dealing with random variables, we need to include the expectation value $E[\cdot]$. A common metric to evaluate point estimators is the squared error, i.e., a quadratic cost function given by $C(p,t) = (p-t)^2$. We can then derive the optimal point estimator associated with this particular cost function:
\begin{align*}
0 &= \frac{\partial E[C(p,t)]}{\partial p}   \\
  &= \frac{\partial \int_{-\infty}^\infty  f(t) (p-t)^2 dt}{\partial p}  \\
 &= 2 \int_{-\infty}^\infty  f(t) (p-t) dt
\end{align*}
Using the normalization of the probability distribution, i.e., $\int_{-\infty}^\infty  f(t) dt = 1$, the equation can be written as 
\begin{equation}
p = \int_{-\infty}^\infty  t f(t) dt
\end{equation}
Therefore, the optimal point estimator associated with a quadratic cost function is the mean of the predicted probability distribution.

In the case of the original newsvendor problem, underage and overage costs are linear. In the special case that they are identical, the cost function is given by the absolute deviation, i.e., $C(p,t) = |p-t|$.
We can derive the optimal point estimator in the following way, using the shorthand notation $\partial_p X$ for $\partial/\partial p X$
\begin{align*}
0 &= \partial_p E[E(p,t)]  \\
&= \partial_p \int_{-\infty}^\infty f(t) |p-t| dt \\
&= \partial_p \left ( \int_{-\infty}^{x_p} f(t) |p-t| dt +  \int_{x_p}^\infty f(t) |p-t| dt \right ), \\
\end{align*}
where we have split the integration at some intermediate point $x_q$. This allows us to resolve the absolute value if we choose $x_q$ such that $t$ is always smaller than $p$ in the first integral and the other way around in the second. Hence:
\begin{equation}
0 = \partial_p \left ( \int_{-\infty}^\infty f(t) (p-t) dt -  \int_{-\infty}^\infty f(t) (p-t) dt \right ) \\
\end{equation}
We then execute the partial derivative with respect to $p$ and obtain:
\begin{equation}
0 = -  \int_{-\infty}^{x_p} f(t) dt + \int_{x_p}^\infty f(t)  dt,
\end{equation}
which we can write as:
\begin{equation}
 \int_{-\infty}^{x_p} f(t) dt = \int_{x_p}^\infty f(t)  dt
\end{equation}
Therefore, the integral from $-\infty$ to the point $x_p$ is the same as from $x_p$ to $\infty$, which means that the point $x_p$ splits the integral in the interval $(-\infty, \infty)$ into two halves. This is the definition of the median, and therefore, a linear cost function with equal over- and underage costs is linked to the median of the predicted PDF as optimal point estimator. If the underage and overage costs differ, the slopes of the two linear segments are no longer the same. This means that the resulting optimal point estimator is no longer the median, but a different quantile. 

\section{Profile Histograms}
\label{sec:profile}

\begin{figure}
\begin{center}
\includegraphics[width=4cm]{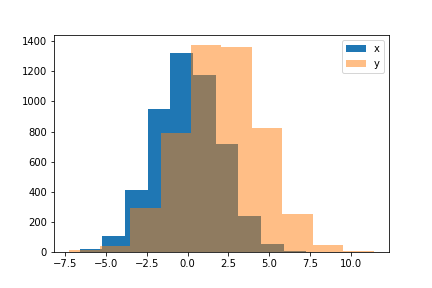}
\includegraphics[width=4cm]{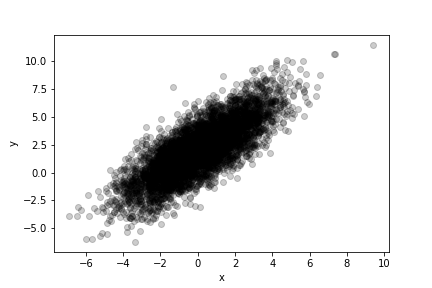}
\caption{\label{fig:scatter} Marginal distributions and scatter plot of variables $X$ and $Y$.}
\end{center}
\end{figure}

In many cases, scatter plots are used to study the behavior of two distributions or sets of data points visually. However, even for a moderate amount of data, this approach quickly becomes difficult. To illustrate this, a sample of $(x,y)$ data points was obtained in the following way: $x$ values by generating 5,000 samples of Gaussian distributed random numbers $X \sim {\cal N}(0.0,2.0)$ and $y$ values via $Y \sim X +  {\cal N}(2.0,1.5)$. \fig \ref{fig:scatter} shows the marginal distributions for $x$ and $y$ as well as a scatter plot of $x$ vs. $y$.

\begin{figure}
\begin{center}
\includegraphics[width=4cm]{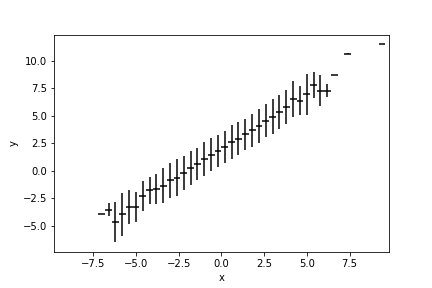}
\caption{\label{fig:profile} Profile histogram of variables $X$ and $Y$.}
\end{center}
\end{figure}

Although the simple linear correlation between $X$ and $Y$ is apparent in the scatter plot, finer details are not visible, and it is easy to imagine that a more complex relationship is difficult to discern. Profile histograms are specifically designed to address this shortcoming. Intuitively, profile histograms are a one-dimensional representation of the two-dimensional scatter plot and are obtained  in the following way: The variable on the $x$ axis is discretized into a suitable range of bins. The exact choice of binning depends on the problem at hand. One can for example choose equidistant bins in the range of the $x$ axis or non-equidistant bins such that each bin contains the same number of observations. Then within each bin of the variable $X$, a location and a dispersion metric are calculated for the variable $Y$. This means that the bin-borders on the $x$ axis are used as constraints on the variable $Y$, and with these conditions applied, for example the sample mean of the selected $y$ values as well as the corresponding standard deviation are calculated. These location and dispersion metrics in each bin of $X$ are used to illustrate the behavior of the variable $Y$ as the values of the variable $X$ change from bin to bin. The resulting profile histogram is shown in \fig \ref{fig:profile}. This one-dimensional representation allows to understand even a complex relationship between two variables visually. Note that due to few data points at the edges of the distributions, the profile histogram is expected to show visual artifacts in the corresponding regions.

\end{document}